%% file: RSPA_Author_tex.tex
\documentclass[openacc]{rsproca_new}

\usepackage[square,numbers]{natbib}
\usepackage{mathrsfs}
\usepackage{amsmath}
\usepackage{algorithm}
\usepackage{algpseudocode}
\usepackage{subfig}
\usepackage[export]{adjustbox}
\usepackage{tabularx}
\input{preamble.macros}

\begin{document}

\title{Physics-informed Dyna-Style Model-Based Deep Reinforcement Learning for Dynamic Control}

\author{
Xin-Yang Liu$^{1}$ and Jian-Xun Wang$^{1}$}

\address{$^{1}$Department of Aerospace \& Mechanical Engineering, College of Engineering, University of Notre Dame, Notre Dame, IN, USA}

\subject{applied mathematics, computational physics, artificial intelligence}


\corres{Jian-Xun Wang\\
\email{jwang33@nd.edu}}

\begin{abstract}
Model-based reinforcement learning (MBRL) is believed to have much higher sample efficiency compared to model-free algorithms by learning a predictive model of the environment. However, the performance of MBRL highly relies on the quality of the learned model, which is usually built in a black-box manner and may have poor predictive accuracy outside of the data distribution. The deficiencies of the learned model may prevent the policy from being fully optimized. Although some uncertainty analysis-based remedies have been proposed to alleviate this issue, model bias still poses a great challenge for MBRL. In this work, we propose to leverage the prior knowledge of underlying physics of the environment, where the governing laws are (partially) known. In particular, we developed a physics-informed MBRL framework, where governing equations and physical constraints are utilized to inform the model learning and policy search. By incorporating the prior information of the environment, the quality of the learned model can be notably improved, while the required interactions with the environment are significantly reduced, leading to better sample efficiency and learning performance. The effectiveness and merit have been demonstrated over a handful of classic control problems, where the environments are governed by canonical ordinary/partial differential equations.
\end{abstract}


\begin{fmtext}
\end{fmtext}


\maketitle

\section{Introduction}
Reinforcement learning (RL) is a class of artificial intelligence (AI) techniques that train an AI agent to learn the optimal control strategy by interacting with the surrounding environment. Over the past few years, with the rapid development of deep learning (DL), deep reinforcement learning (DRL) techniques have been witnessing tremendous success in a variety of applications. In particular, DRL has demonstrated superhuman performance at playing Go~\cite{silver2017mastering,silver2018general} and Atari games from pixels~\cite{mnih2013playing}. Most recently, there has been growing interest in applying DRL for dynamic control of complex physical systems, e.g., laminar/turbulent flows~\cite{rabault2019artificial,ghraieb2021single,ren2021applying,fan2020reinforcement,bucci2019control}, active matter~\cite{falk2021learning}, fish swimmers~\cite{gustavsson2017finding,verma2018efficient,zhu2021numerical}, unmanned aerial vehicles (UAV)~\cite{hwangbo2017control,wada2021unmanned,deng2021event}, and robotics~\cite{bhagat2019deep,nagabandi2020deep,li2020learning}.     

In general, most state-of-the-art RL agents learn the desired tasks by gathering experience directly from the environment. Namely, the optimal action strategy is derived by interacting with the real physical system in a trial-and-error manner. This class of RL methods is known as \emph{model-free reinforcement learning} (MFRL). Due to the ease of implementation and no need for prior knowledge of the dynamic transitions, MFRL has been widely applied to many tasks, mainly in playing computer games~\cite{botvinick2019reinforcement,mnih2013playing,hessel2018rainbow}. Despite their popularity, MFRL methods usually have \emph{low sample efficiency}, i.e., requiring a massive amount of interactions with the environment. Although the low sample efficiency and slow convergence rate might be acceptable for training an agent in gaming applications since the environment interactions are nearly costless, these shortcomings will significantly limit the DRL applications for dynamic control of physical/mechanical systems (e.g., flights or robotics). First, real-world interactions of a mechanical system can be very expensive and time-consuming. For instance, considering a flow control problem with plasma actuators, it is very costly or even infeasible to train a controller by conducting a huge amount of wind tunnel experiments with enormous control trials, which is unlike training an AI game agent that can be done by playing the computer games for a vast amount of times (episodes). Second, the mechanical systems can be easily worn out from extensive action trials, and thus the exploration of optimal control strategy in MFRL is highly restricted to avoid possible damage to the system in real-world industry settings. Although the off-policy MFRL algorithms with a replay buffer (e.g., deep Q-Networks~\cite{mnih2013playing} and their actor-critic extensions~\cite{lillicrap2015continuous,fujimoto2018addressing,haarnoja2018soft}) can better utilize historical data than on-policy MFRL algorithms (e.g., trust region policy optimization~\cite{schulman2015trust} and proximal policy optimization~\cite{schulman2017proximal}), the data efficiency is still far from sufficient.
 
One way to improve data efficiency is to augment the data collected from real-world interactions with a learned transition model. This is the general idea of the other class of DRL algorithms: \emph{Model-based reinforcement learning} (MBRL)~\cite{moerland2020model,plaat2020model}. Using a learned model to reason about the future can avoid the irreversible consequence of trial-and-error in the real environment and has great potential to significantly improve data efficiency, which is thus more appealing in applications of complex mechanical systems. In addition, the learned transition model is independent of rewards and thus can be transferred to other control problems in the same/similar environments. Many existing MBRL methods rely on simple function approximators, such as Gaussian process (GP), linear models, and Gaussian mixture models~\cite{deisenroth2011pilco,tassa2012synthesis,levine2013guided}. However, the limited expressibility of the simple models prevents them from handling high-dimensional problems with complex dynamic transitions. Thanks to the rapid developments of deep learning, more and more complex high-dimensional function approximators based on neural networks have been applied to design more powerful MBRL algorithms. For example, Racani{\`e}re et al.~\cite{racaniere2017imagination} proposed a novel MBRL framework, Imagination-Augmented Agent (I2A), where the environment model is constructed by a recurrent network architecture for generating imagined trajectories to inform agent's decisions. Kaiser et al.~\cite{Kaiser2020Model} presented a complete MBRL method (SimPLe) using a convolutional neural network to successfully solve Atari games with significantly fewer interactions than MFRL methods. Hafner et al.~\cite{hafner2019learning} developed the Deep Planning Network (PlaNet) that learns the latent dynamics of the environment directly from images using a variational autoencoder and a recurrent latent network. The effectiveness of PlaNet has been demonstrated by successfully solving a number of continuous control tasks from pixels. Hafner et al.~\cite{Hafner2020Dream} further extended the PlaNet by developing a novel actor-critic based MBRL method (Dreamer), which learns long-horizon behaviors from images purely by latent imagination. Dreamer has been evaluated on the DeepMind Control Suite and outperforms most state-of-the-art MBRL and MFRL algorithms in every aspect. 

Despite the great promise, most commonly-used MBRL approaches suffer from model inaccuracy (i.e., model bias), preventing them from matching the success of their model-free counterparts~\cite{moerland2020model}. This is particularly true when it comes to learning complex dynamics with high-capacity models (e.g., deep neural networks), which are prone to overfitting in data-sparse and out-of-sample regimes~\cite{plaat2020model}. In particular, the model bias can be significantly exacerbated for predicting long rollout horizons because of the ``compound error'' effect. To mitigate this issue, rather than learning the transition deterministically, people built the dynamic models in a probabilistic manner, where the unknown model bias is treated as the epistemic uncertainty (i.e., model-form uncertainty) and can be modeled in several different ways. For example, Depeweg et al.~\cite{depeweg2016learning} employed Bayesian neural networks (BNNs) to learn the probabilistic dynamic transition and update the policy over an ensemble of models sampled from the trained BNNs. Kurutach et al.~\cite{kurutach2018modelensemble} proposed to use an ensemble of models to estimate the model-form uncertainty and regularize the trust region policy optimization (TRPO). Nonetheless, the model-form uncertainty is notoriously difficult to quantify, especially for black-box deep learning models~\cite{abdar2021review,wang2018propagation}. Most recently, a more promising strategy known as physics-informed deep learning (PIDL) has attracted increasing attention in the scientific machine learning (SciML) community, aiming to leverage both the advantages of deep learning and prior knowledge of underlying physics to enable data-scarce learning. Instead of learning solely from labeled data, the model training process is also guided by physics laws and knowledge, which could provide rigorous constraints to the model output, alleviate overfitting issues, and improve the robustness of the trained model in data-scarce and out-of-sample regimes. This idea has been recently explored for solving PDEs or modeling complex physical systems. For example, researchers have incorporated physical constraints (e.g., realizability, symmetry, invariance) into SciML models to develop physics-informed, data-driven turbulence models~\cite{wang2017physics,ling2016reynolds,duraisamy2019turbulence}. People have also utilized governing equations of the physical systems to inform or directly train deep neural networks, i.e., physics-informed neural networks (PINNs)~\cite{raissi2019physics}, which has been demonstrated in many scientific and engineering applications~\cite{sun2020surrogate,gao2020super,arzani2021uncovering,sahli2020physics,rao2021physics}.

In this work, we leverage the idea of PIDL and propose Physics-informed Model-Based Reinforcement Learning (PiMBRL), an innovative MBRL framework for complex dynamic control that incorporates the physical laws/constraints of the system to alleviate the issue of model bias, reduce the real-world interactions, and significantly improve the data efficiency. Specifically, a novel autoencoding-based recurrent network architecture is constructed to learn the dynamic transition in the Dyna-style MBRL framework~\cite{sutton2018reinforcement}, which is a commonly-used MBRL formulation. The governing physics of the environment are assumed to be known and are utilized to inform the model learning and RL agent optimization. State-of-the-art Off-policy actor-critic (AC) algorithms, e.g., Twin Delayed Deep Deterministic Policy Gradients (TD3)~\cite{fujimoto2018addressing}, are used for value/policy optimization, 
We have demonstrated the effectiveness and merit of the proposed PiMBRL on a few classic dynamic control problems, where the environments are governed by canonical ordinary/partial differential equations (ODEs/PDEs), including cart-pole, pendulum, viscous fluid dynamics governed by Burgers' equation, and chaotic/turbulent dynamics governed by Kuramoto-Sivashinsky (KS) Equation. The performance of the proposed PiMBRL algorithms is compared with their MBRL and MFRL counterparts, and significant improvements in terms of sample efficiency and model accuracy are observed. The novel contributions of this work are summarized as follows: (a) we propose a physics-informed model-based RL framework based on a novel encoder-decoder recurrent network architecture; (b) embed the physics of the environment into the MBRL using discretized PIDL formulation~\cite{gao2021phygeonet}; (c) demonstrate the effectiveness of proposed methods on a variety of dynamic control problems, particularly including nonlinear spatiotemporal chaotic systems, e.g., the KS equation, which exhibits a wide range of dynamics from the steady to chaotic/turbulent regimes, shedding lights on developing controllers for more challenging fluid systems governed by Navier-Stokes equations; (d) compare the proposed method with state-of-the-art MBRL and MFRL in terms of accuracy and sample complexity. This work is the first attempt to use physical laws to inform the MBRL agent optimization to the best of the authors' knowledge.

The rest of the paper is organized as follows. The background of MFRL/MBRL and our proposed PiMBRL algorithms are introduced in Section~\ref{sec:meth}. Results of numerical experiments on classic dynamic control problems are presented and discussed in Section~\ref{sec:result}. The influence of model rollout length and model accuracy threshold is further discussed in Section~\ref{sec:discuss}. Finally, Section~\ref{sec:conclude} concludes the paper. 


\section{Methodology}
\label{sec:meth}
\subsection{Problem formulation and background}
We consider dynamical systems equipped with localized control inputs (i.e., actuators),
\begin{equation}
    \frac{d \ubm}{dt} = \mathscr{F}(\ubm, \abm; \mubold),
\end{equation}
where $\ubm(\xbm, t) \in \mathbb{R}^{d_u}$ denotes the state variable of the system in the spatial domain $\Omega$ and temporal domain $t \in [0, T]$, $\abm(\xbm, t) \in \mathbb{R}^{d_a}$ represents the action variable (i.e., control inputs), and $\mathscr{F}(\cdot)$ is a nonlinear differential operator parameterized by $\mubold$. In many cases, the systems can be assumed to possess the Markov property, referred to as Markov Decision Processes (MDP). The discrete form can be written as, 
\begin{equation}
    \ubm_{t+1} = \mathcal{F}(\ubm_{t}, \abm_{t}; \mubold),
\end{equation}
where the state $\ubm_{t+1}$ at next time $t+1$ only depends on the state $\ubm_{t}$ and action $\abm_t$ at current time step $t$, and the time-invariant transition dynamics $\mathcal{F}$ of the environment is a nonlinear differential functional. In the optimal control problem, the goal is to find a series of action signals (a.k.a., policy $\pi$) that maximizes the expected return $R(\pi)$,
\begin{equation}
    R(\pi) = \int_0^T{\mathop{\mathbb{E}}_{\ubm_t \sim \pi_t}\bigg[r(\ubm_t)\bigg]},
\end{equation}
where $r(\ubm_t)$ denotes the reward function of the state at time $t$, which is a signal to assess the control agent locally. This optimal control problem can be solved by deep reinforcement learning (DRL), either in a model-free or model-based manner. 

\subsubsection{Value function, policy function, and Bellman equation}
Before putting forth the proposed DRL algorithms, we introduce several important concepts in DRL, including value \& policy functions and Bellman equation. Value functions are functions of a state (or a state-action pair) that estimate the total return starting from that particular state (state-action pair). Value function $v(\ubm)$ of a state $\ubm$ is known as state-value function, while value function $q(\ubm, \abm)$ of a state-action pair $(\ubm, \abm)$ is known as action-value function. The state-value and action-value functions are formally defined as,
\begin{subequations}
\begin{align}
v(\ubm) &\doteq \mathbb{E}\left[\sum_{k = 1}^\infty{\gamma^k r(\ubm_{t + k})} \Bigg| \ubm_t = \ubm \right], \\
q(\ubm, \abm) &\doteq \mathbb{E}\left[\sum_{k = 1}^\infty{\gamma^k r(\ubm_{t + k})} \Bigg| \ubm_t = \ubm, \abm_t = \abm \right],
\end{align}
\label{eq:valuefunction}
\end{subequations}
where $\gamma \leq 1$ is the discount rate. A policy function $\pi$ maps states to actions (or probabilities of actions). Namely, a policy function $\pi$ can be defined either as a deterministic function $\pi(\ubm) = \abm$ or a probability measure $\pi(\abm|\ubm)$. Due to the nature of MDP, value functions can be estimated recursively based on the Bellman equations~\cite{sutton2018reinforcement},
\begin{subequations}
\begin{align}
v(\ubm_t) &= \mathop{\mathbb{E}_\pi}_{\ubm_{t+1} \sim \mathcal{P}}\bigg[ r(\ubm_t, \abm_t) + \gamma v(\ubm_{t+1})\bigg], \\
q(\ubm_t, \abm_t) &= \mathop{\mathbb{E}}_{\ubm_{t+1} \sim \mathcal{P}} \bigg[ r(\ubm_t, \abm_t) + \gamma\mathop{\mathbb{E}}_{\abm_{t+1} \sim \pi}q(\ubm_{t+1}, \abm_{t+1})\bigg],
\end{align}
\label{eq:bellman}
\end{subequations}
where $\mathcal{P} = \mathcal{P}(\ubm_{t+1}|\ubm_t, \abm_t)$ is the transition probability, describing the dynamics of the environment.

\subsubsection{Model-free and model-based reinforcement learning}
\begin{figure}[!htp]
\centering\includegraphics[width=1\textwidth]{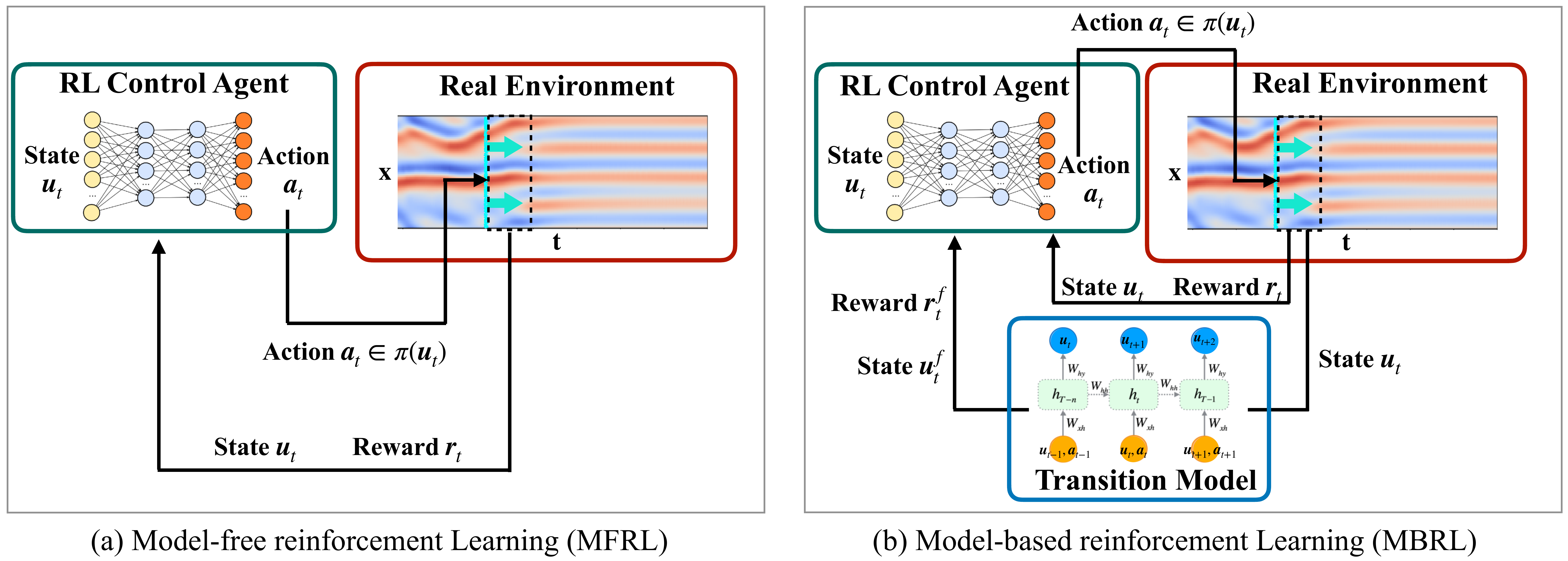}
\caption{Schematics of model-free reinforcement learning (MFRL) and model-based reinforcement learning (MBRL).}
\label{fig:overview}
\end{figure}

As mentioned above, RL aims to find a series of actions (i.e., optimal policy) that maximize the total return by estimating the value and/or policy function based on the Bellman equation. Depending on whether or not learning and using a model of the transition dynamics of the environment, RL can be classified into two categories: model-free reinforcement learning (MFRL) and model-based reinforcement learning (MBRL). In MFRL, the optimization process is conducted by repeatedly interacting with the environment with a trial-and-error search, and the model of the environment is not required (see Fig.~\ref{fig:overview}a). Namely, the state dynamics of the environment are (partially) observed as exploring different policy strategies, and the best policy will be identified after massive trials. MBRL, on the other hand, leverages a model $\tilde{\mathcal{F}}$ that approximates the real environment $\mathcal{F}$ and predicts the dynamic transition ($\tilde{\mathcal{F}}: \ubm_t, \abm_t \to \ubm_{t+1}$), which can be learned from the interactions with the real environment. The RL agent is then optimized by the interactions not only with the real environment but also with the virtual environment constructed by the model (see Fig.~\ref{fig:overview}b). The learned model $\tilde{\mathcal{F}}$ can be utilized for planning with its gradient information (e.g., SVG~\cite{NIPS2015_14851003}, GPS~\cite{levine2014learning}) or synthesizing imagined samples to augment real samples for better sample efficiency. The latter is known as the Dyna-like MBRL~\cite{sutton2018reinforcement,luo2018algorithmic,kurutach2018modelensemble,Kaiser2020Model} that can directly leverage cutting-edge MBRL algorithms.

The algorithms of the DRL agent optimization can be grouped into three classes: (i) actor-only, (2) critic-only, and (3) actor-critic methods. Actor-only (i.e., policy-gradient) methods directly optimize the policy function $\tilde{\pi}(\ubm;\boldsymbol{\theta}_\pi)$, which is often parameterized by a deep neural network by calculating the policy gradient with respect to network parameters $\boldsymbol{\theta}_\pi$. The optimization can be solved based on stochastic gradient descent (SGD)~\cite{amari1967theory} or its variants~\cite{konevcny2013semi,johnson2013accelerating,defazio2014saga,schmidt2017minimizing}. Examples of policy-gradient methods include REINFORCE~\cite{williams1992simple}, TRPO~\cite{schulman2015trust}, PPO~\cite{schulman2017proximal}. Critic-only (value-based) methods are a family of RL algorithms that learn an DNN-approximated value-action function $\tilde{q}(\ubm, \abm; \boldsymbol{\theta}_q)$ based on the optimal Bellman equation. Examples include Q-Learning~\cite{watkins1992q}, DQN~\cite{mnih2013playing}, Dueling DQN~\cite{wang2016dueling}, etc. Since the optimization in critic-only methods is always performed off-policy, they are more sample-efficient than the actor-only methods that are often on-policy. However, since critic-only methods optimize the policy indirectly, they are less stable compared to the actor-only methods. The actor-critic (AC) methods learn a value function to support the policy gradient optimization, and thus the AC family combines the strengths of both actor and critic methods. As such, AC-based methods will be used in the proposed PiMBRL for value/policy optimization.   

\subsection{Physics-informed model-based reinforcement learning}
We propose a physics-informed model-based reinforcement learning (PiMBRL) framework, where the physics knowledge (e.g., conservation laws, governing equations, and boundary conditions) of the environment is incorporated to inform the model learning and RL optimization. In this work, we focus on the Dyna-style MBRL formulation with the off-policy AC-based optimization. The proposed framework will retain the generality and optimality of model-free AC-based DRL methods, while significantly reducing the real-world interactions by learning a reliable environment model based on physics-informed discrete learning. 
\begin{figure}[!htp]
\centering\includegraphics[width=0.75\textwidth]{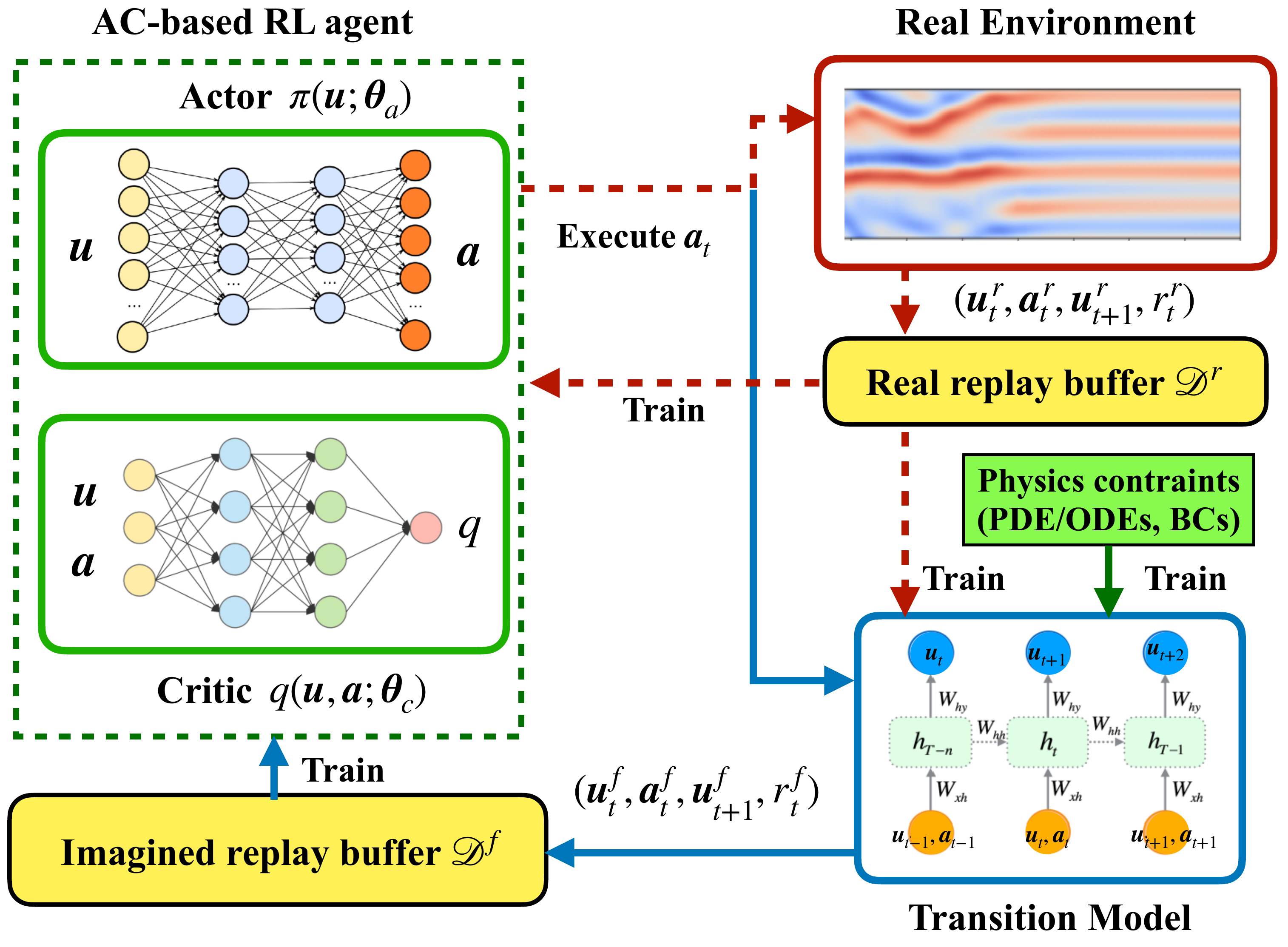}
\caption{Schematics of Dyna-style physics-informed model-based actor-critic algorithm.}
\label{fig:pimbrl}
\end{figure}
Specifically, an AC-based RL agent will be initialized and starts interacting with the real environment. State-action data pairs and corresponding rewards $(\ubm_t^r, \abm_t^r, \ubm_{t+1}^r, r_t^r)$ are iteratively collected from the real environment and saved into the real replay buffer $\mathscr{D}^r$. These real samples are used to train the actor-critic agent and the transition model simultaneously. The model is constructed by an auto-encoding recurrent network, where boundary conditions (BCs) of the system are strictly encoded, and the governing physics are imposed softly by minimizing the violation of the conservation laws. Synthetic samples $(\ubm_t^f, \abm_t^f, \ubm_{t+1}^f, r_t^f)$ are generated by the model and are collected into the imagined replay buffer $\mathscr{D}^f$, which will be leveraged to augment the real samples for the RL update. The model training, data generation, environment interaction, and RL agent optimization are conducted iteratively in an online manner. The overall schematic of the proposed PiMBRL is shown in Fig.~\ref{fig:pimbrl}, and the detailed algorithm is given by Algorithm~\ref{algo1}. More details of AC optimization and physics-informed model construction will be elaborated in the following subsections.
\begin{algorithm}
\caption{Physics-informed Model-Based Reinforcement Learning (PiMBRL)}\label{algo1}
\begin{algorithmic}[1]
\State Randomly initialize policy (actor) network $\pi(\ubm;\boldsymbol{\theta}_{\pi})$, value (critic) network(s) $q(\ubm, \abm;\boldsymbol{\theta}_{q})$, transition model $\tilde{\mathcal{F}}(\ubm, \abm; \boldsymbol{\theta}_F)$, and replay buffers $\mathscr{D}^r, \mathscr{D}^f$ for real and fictitious environments.
\State Randomly initialize state $\ubm_0$ (or observed state $\ubm^o_0$), reward $r_0$, and done signal $d_0$ for real environment.
\For {episode$=1, M$}
    \For {$i=0, T$}
        \State Execute action $\abm_i = \pi(\ubm;\boldsymbol{\theta}_{\pi})$ in the real environment $\mathcal{F}$;
        \State Save new data pair $(\ubm^o_i,\abm_i,\ubm^o_{i+1},r_{i},d_{i})$ to the real buffer $\mathscr{D}^r$;
        \If {episode ends}
            Reset the environment $\mathcal{F}$;
        \EndIf
    \EndFor
    \If{sufficient ($n_{s_M}$) state-action pairs stored in the real buffer $\mathscr{D}^r$}
        \State Sample a batch of real state-action pairs, $\{(\ubm^o_i,\abm_i,\ubm^o_{i+1})\}_{i=1}^{n_{b_r}}$, from $\mathscr{D}^r$
        \State Update the transition model $\tilde{\mathcal{F}}(\ubm, \abm; \boldsymbol{\theta}_F)$ using the data loss $L_D$ on the batch $\mathscr{D}^r$;
    \EndIf
    
    \If {transition model meets the accuracy threshold (data loss $L_D < \lambda$)}
        \For {model prediction length$=1, l_M$}
            \State Sample a batch of states $\{(\ubm_{j+1})\} $ from $\{\mathscr{D}^r,\mathscr{D}^f\}$
            \State Execute actions $\{\abm_{j+1}=\pi(\ubm_{j+1};\boldsymbol{\theta}_{\pi})\}^{n_{b_f}}_{j=1}$ in transition model $\tilde{\mathcal{F}}$;
            \State Save new data pairs $\{(\ubm^o_{j+1},\abm_{j+1},\ubm^o_{j+2},r_{j+2},d_{j+2})\}^{n_{b_f}}_{j=1}$ to buffer $\mathscr{D}^f$;
        \EndFor
    \EndIf
    \If{enough ($n_{s_R}$) state-action pairs stored in $\{\mathscr{D}^r,\mathscr{D}^f\}$}
        \State Sample a batch of state-action pairs $\{(\ubm^o_k,\abm_k,\ubm^o_{k+1})\}$ from the fake buffer $\mathscr{D}^f$
        \State Update model $\tilde{\mathcal{F}}$ according to physical loss $L_E$ on sampled state-action pairs;
    \EndIf
    \State Sample a batch of $ \{(\ubm^o_l,\abm_l,\ubm^o_{l+1},r_{l},d_{l})\}$ from the augmented buffer $\{\mathscr{D}^r,\mathscr{D}^f\}$
    \State Update policy network $\pi(\ubm;\boldsymbol{\theta}_{\pi})$ and value network $q(\ubm, \abm;\boldsymbol{\theta}_{q})$ on the sampled state-action pairs using off-policy algorithms\footnotemark (see Algorithm \ref{integ-TD3} in Appendix).
\EndFor
\end{algorithmic}
\end{algorithm}
\footnotetext{In this paper, TD3 is used as a demonstration (details of the TD3 is given in Algorithm \ref{integ-TD3}), but other off-policy algorithms such as Deep Deterministic Policy Gradient (DDPG) and Soft Actor-Critic (SAC) are also applicable.}

\subsubsection{Dyna-style model-based actor-critic optimization}
We consider a generic Dyna-style model-based actor-critic (AC) optimization algorithm, which can be easily adapted to any state-of-the-art off-policy AC methods, e.g., DDPG, TD3, or SAC. As mentioned above, in the Dyna-style formulation, model is used to augment real samples, and thus model-free AC-based optimization can be directly leveraged. For a AC-based RL agent, two neural networks $\tilde{\pi}(\ubm;\boldsymbol{\theta}_{\pi})$ and $\tilde{q}(\ubm, \abm; \boldsymbol{\theta}_{q})$ are constructed to represent the policy and value functions, respectively. Based on samples from both the real environment and virtual environment simulated by the model, the policy network $\tilde{\pi}(\ubm;\boldsymbol{\theta}_{\pi})$ is iteratively updated by,
\begin{equation}
    \boldsymbol{\theta}_{\pi}^{k+1} = \boldsymbol{\theta}_{\pi}^{k} + \alpha_\pi \nabla_{\boldsymbol{\theta}_{\pi}}J(\boldsymbol{\theta}_{\pi}^{k}),
\end{equation}
where $\alpha_\pi$ is the learning rate and $\nabla_{\boldsymbol{\theta}_{\pi}}J(\boldsymbol{\theta}_{\pi})$ represents the policy gradient with respect to actor network parameters $\boldsymbol{\theta}_{\pi}$, which can be calculated based on the critic network,
\begin{equation}
    \nabla_{\boldsymbol{\theta}_{\pi}}J(\boldsymbol{\theta}_{\pi}^{k}) = \mathbb{E}\left[ \sum_{t=0}^{T} \nabla_{\boldsymbol{\theta}_{\pi}} \log \tilde{\pi}(\ubm_t; \boldsymbol{\theta}_{\pi}^{k}) \cdot \tilde{q}(\ubm_t, \abm_t; \boldsymbol{\theta}_{q}^{k}) \right].
\end{equation}
The critic network $\tilde{q}(\ubm, \abm; \boldsymbol{\theta}_{q})$ is optimized by minimizing the temporal difference (TD)-based loss function,
\begin{equation}
    \boldsymbol{\theta}_{q}^* = \underset{\boldsymbol{\theta}_{q}}{\arg\min} \|q'_{t} - \tilde{q}(\ubm_t, \abm_t; \boldsymbol{\theta}_{q})\|_{L_2}, 
\end{equation}
where $\|\cdot\|_{L_2}$ represent L2 norm and $q'_{t}$ is estimated based on the optimal Bellman equation,
\begin{equation}
    q'_{t} = r_t + \gamma \tilde{q}\bigg(\ubm_{t+1}, \tilde{\pi}(\ubm_{t+1};\boldsymbol{\theta}_{\pi}); \boldsymbol{\theta}_{q}\bigg).
\end{equation}

\subsubsection{Physics-informed Learning architecture for transition dynamics}
\begin{figure}[htp!]
    \centering
    \includegraphics[width=0.9\textwidth]{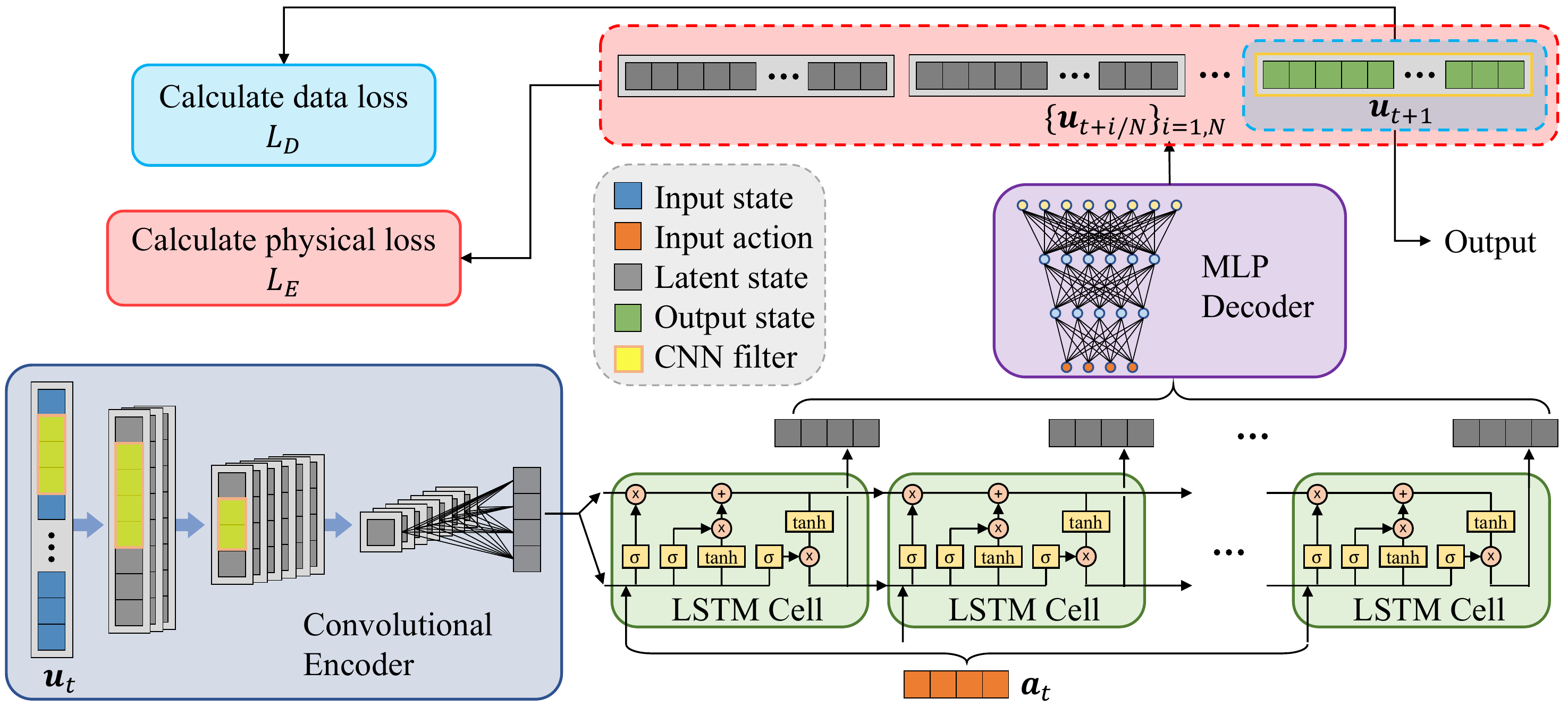}
    \caption{Schematics of LSTM-based neural network architecture for transition model}
    \label{fig:nn_model}
\end{figure}

We develop a physics-informed recurrent neural network to learn the dynamics transition, aiming to map the current states and actions to the states at the next control step ($\tilde{\mathcal{F}}: \ubm_t, \abm_t \to \ubm_{t+1}$). To better capture the spatiotemporal dependencies, a convolutional encoder, multi-layer perceptron (MLP) decoder, and long-short term memory (LSTM) blocks are utilized to build the learning architecture. As shown in Fig.~\ref{fig:nn_model}, the high-dimensional state vector ($\ubm_t$) at the current control step is encoded into the latent space by a convolutional encoder. Together with the input actions ($\abm_t$), the latent state vector is fed into the LSTM-based transition network, which outputs latent intermediate transition states between the two control steps. After a multi-layer perceptron (MLP) decoder, the latent outputs are decoded to the full-order physical states.         
The network is trained based on both data and physics constraints. The data loss $L_D$ is defined by the mismatch between the model prediction $\tilde{\ubm}_{t+1}$ and labeled data $\ubm_{t+1}$ obtained from interactions with the real environment,
\begin{equation}
    L_D = \frac{1}{n_b}\sum_{j=1}^{n_b} \,\left\|\tilde{\ubm}_{t+1}^{(j)} - \ubm_{t+1}^{(j)}\right\|_{L_2},
\end{equation}
where $n_b$ is the batch size. The physical loss $L_E$ is constructed based on the conservation laws of the system in their discretized form. The residuals of the governing equations is minimized on multiple discrete spatiotemporal snapshots. To this end, the network outputs include the states at $N$ intermediate time steps ($\{\ubm_{t+i/N}\}_{i=1,N}$) as well as the state at the next control step. The physical loss is then obtained by taking the averaged residuals of the governing equations, 
\begin{equation}
    L_E = \frac{1}{N}\sum_{i=1}^{N}\sum_{j=1}^{n_b} \,\left\|\frac{d \tilde{\ubm}_{t+i/N}^{(j)}}{d t} - \mathcal{F}(\tilde{\ubm}_{t+i/N}^{(j)},\, \abm_{t+i/N})^{n_b}\right\|_{L_2},
\end{equation}
where spatial derivatives in $\mathcal{F}$ are computed using high-order finite-difference-based spatial filtering, and temporal derivatives $\frac{d \ubm_{t+i/N}}{d t}$ are approximated by forward Euler method.

\section{Results}
\label{sec:result}
In this section, we test the proposed PiMBRL on a number of classic control problems that are governed by ODEs or PDEs. The performance is compared against the baseline model-free and model-based counterparts. The standard model-free TD3 algorithm is used as the MFRL baseline (see Algorithm~\ref{MF-TD3} in Appendix), while the purely data-driven dyna-like model-based TD3 is deemed as the MBRL baseline. The hyperparameters used in the following experiments are summarized by Tables~\ref{tab:TD3-hp} and~\ref{tab:MBRL_hp} in Appendix. 

\subsection{ODE governed environments}\label{sec:ode}
We first evaluate PiMBRL on two classic dynamic control benchmarks, Cart-Pole and Pendulum, which are available in the OpenAI Gym environment. The physics of both systems are known, which can be described by a set of ODEs. Since there is no spatial dependence and the dimension of the system is low, we use a low-capacity two-layer MLP with 256 neurons per layer to directly learning the transition dynamics $\ubm_{t+1}=\tilde{\mathcal{F}}(\ubm_{t},\abm_{t})$ and model rollout length ($l_M$) is set equal to the trajectory length.   

\subsubsection{Cart-Pole}
\begin{figure}[htp!]
    \centering
    \subfloat[Cart-Pole]{\includegraphics[width=0.32\textwidth]{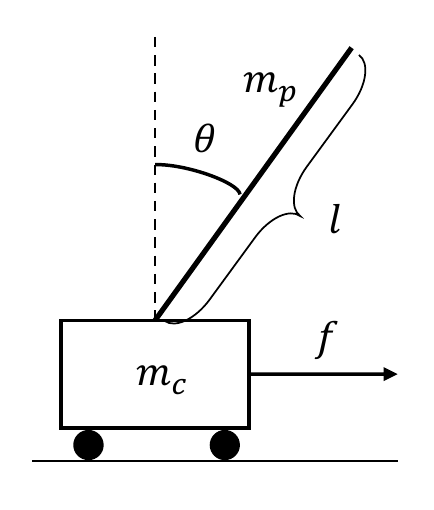}}\,
    \subfloat[RL Performance]{\includegraphics[width=0.5\textwidth]{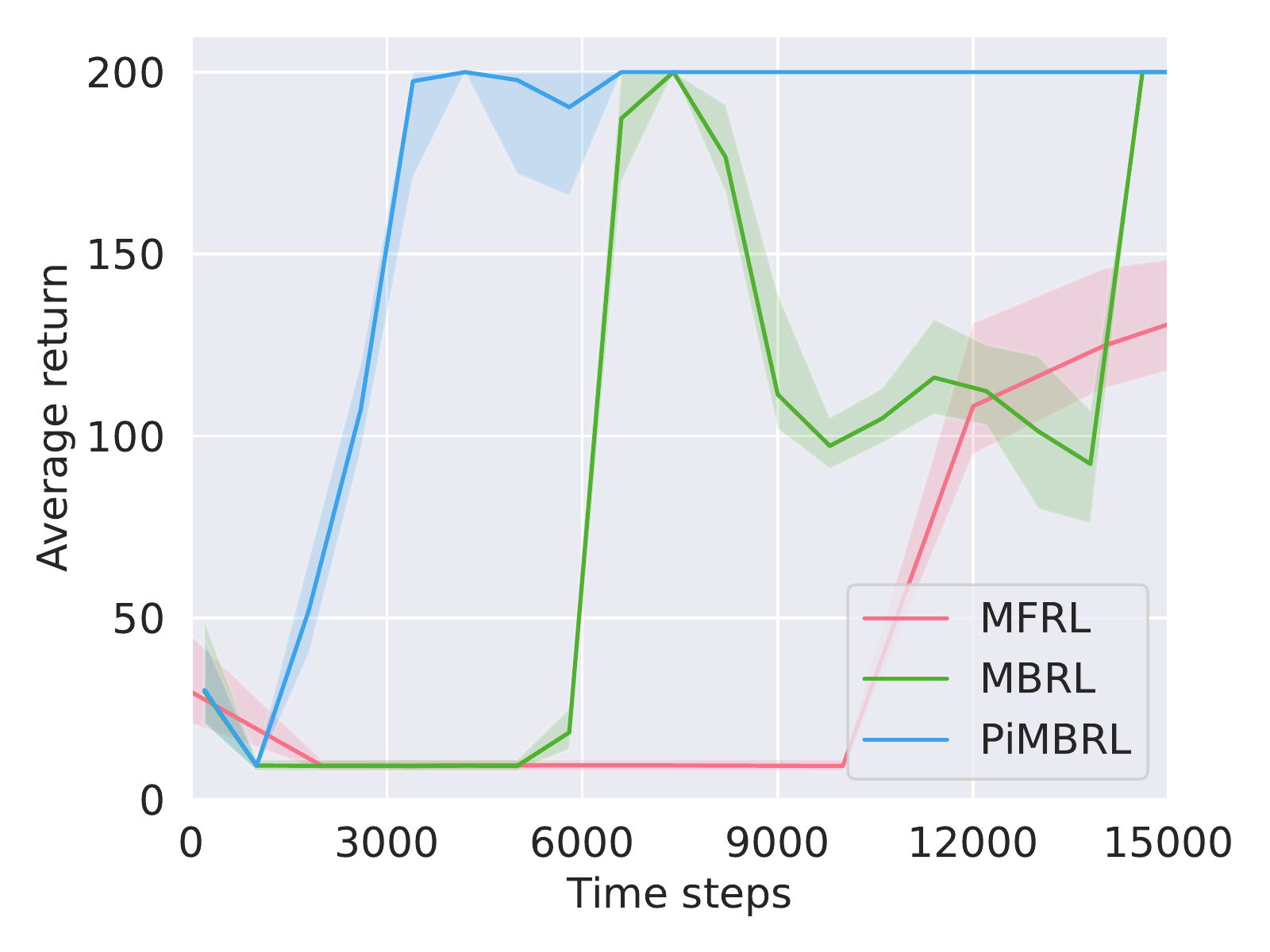}}
    \caption{(a) Schematic diagram of the Cart-Pole environment. (b) Performance curve of PiMBRL versus standard model-free TD3 (MFRL) and dyna-like model-based TD3 (MBRL) in Cart-Pole environment. The solid lines indicate averaged returns of 100 randomly selected test episodes, while the shaded area represents the return distribution of all test samples.}
    \label{fig:cartpole}
\end{figure}

We start with the Cart-Pole benchmark problem (i.e., ``CartPole-v0'' environment provided in OpenAi gym), where a cart moves along a frictionless track with a pole attached to the top of it via an unactuated joint, as shown in Fig.\ref{fig:cartpole} (a). The control goal is to keep the pole from falling over by acting a horizontal force on the cart. The reward is $+1$ for each time step as long as the pole is upright and the cart remains in a certain region. The physics of this system is governed by,
\begin{equation}
    \begin{cases}
    \displaystyle\ddot{x} = \frac{f + m_p \dot{\theta}^2 \sin{\theta}- m_p l \ddot{\theta}}{m_p + m_c}  \\
    \displaystyle\ddot{\theta} = \frac{g\sin{\theta}-\cos{\theta}\displaystyle\frac{f + m_p \dot{\theta}^2 \sin{\theta}
}{m_p + m_c}}{l\left(\displaystyle\frac{4}{3}-\frac{m_p\cos^2{\theta}}{m_c+m_p}\right)},
    \end{cases}
    \label{eq:cartpole}
\end{equation}
where $x$ is the spatial coordinate of the cart, $\theta$ represents the angle of the pole from vertical, and $f$ is the force that the RL agent applies on the cart. $m_c, m_p$ are the mass of the cart and pole, respectively. One episode is considered to be ended if the pole deviates too much from vertical position (i.e., $|\theta| > \pi/12$) or the cart leaves the designated area (i.e., $|x| >2.4$) or one episode has more than 200 control steps. The state observation of this environment is a four-dimensional vector, $\ubm=(x,\dot{x},\theta,\dot{\theta})$, while the action space is discrete, consisting of two valid values $\{-10,10\}$. Each episode begins at a random state $\ubm_0 = (x_0,\dot{x}_0,\theta_0,\dot{\theta}_0)$.

Figure \ref{fig:cartpole} (b) compares the performance curves of the MFRL, MBRL, and PiMBRL. The proposed PiMBRL reaches the total return of 200 only after about 3000 time steps in the real environment, while the vanilla dyna-like MBRL counterpart needs much longer to achieve so and its performance is not stable as well. For the MFRL counterpart, the total return is still below 120 even after 15000 time steps. Although both MBRL and MFRL are able to achieve the same performance with sufficient time steps, PiMBRL only uses about 45.2\% and 9.7\% time steps needed by its MBRL and MFRL counterparts, respectively. Therefore, to achieve the same level of performance, PiMBRL can significantly reduce the required number of interactions with the real environment, compared to the original model-free TD3 (i.e., MFRL) and dyna-like model-based TD3 (i.e., MBRL).

\subsubsection{Pendulum}
\begin{figure}[htp!]
    \centering
    \subfloat[Pendulum]{\includegraphics[width=0.28\textwidth,valign=c]{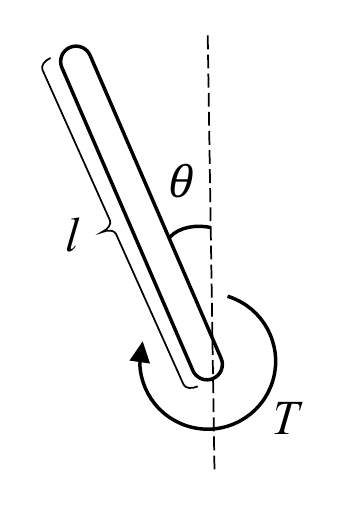}}\quad \quad \qquad
    \subfloat[RL performance]{\includegraphics[width=0.5\textwidth,valign=c]{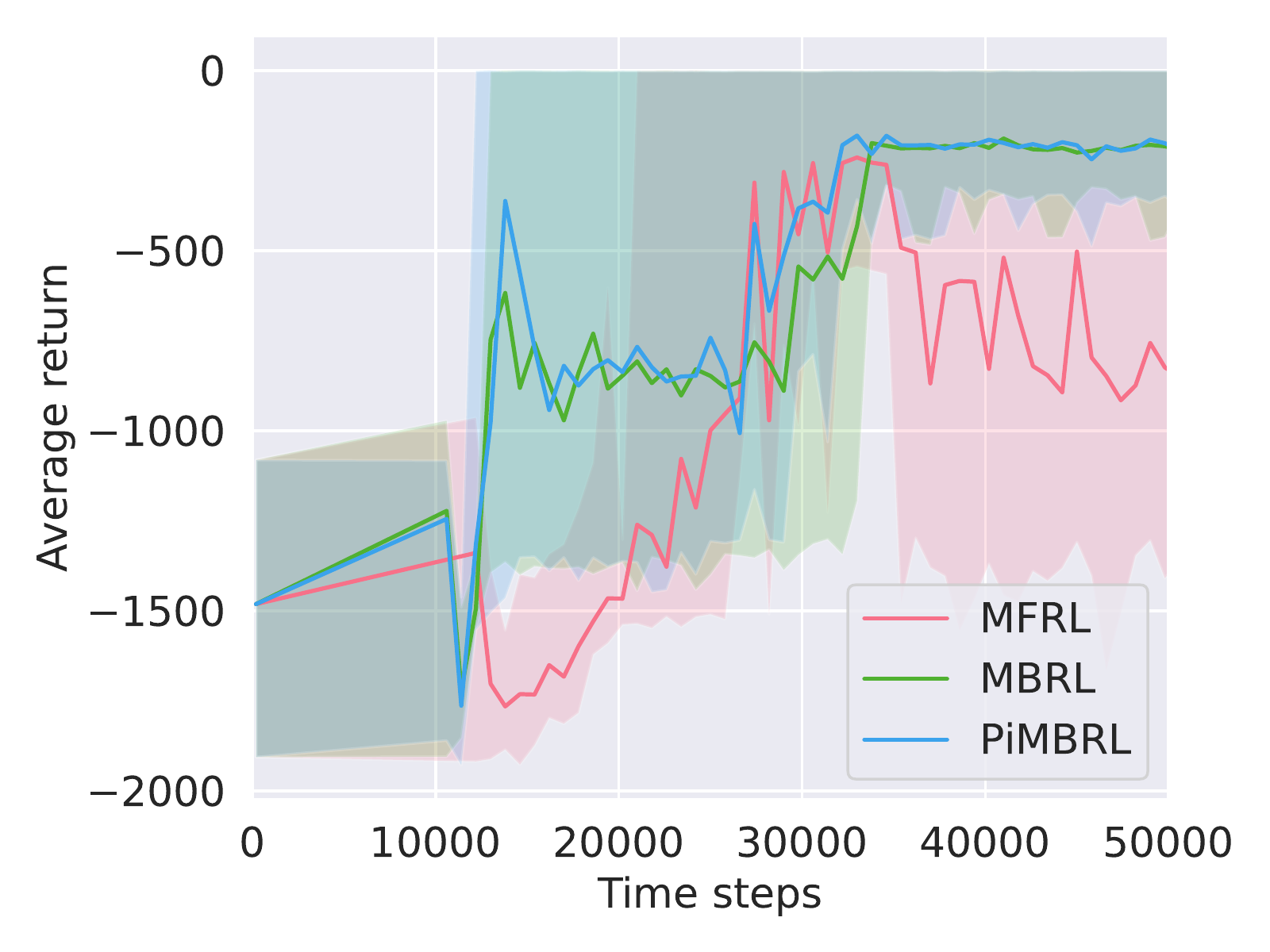}}\\
    
    \caption{(a) Schematic diagram of pendulum environment. (b) Performance curve of PiMBRL versus standard model-free TD3 (MFRL) and dyna-like model-based TD3 (MBRL) in the pendulum environment. The solid lines indicate averaged returns of 100 randomly selected test episodes, while the shaded area represents the return distribution of all test samples.}
    \label{fig:pen}
\end{figure}

The second test case is the Pendulum-v0 available in the OpenAi gym. In this environment (see Fig. \ref{fig:pen} (a) ), one end of the pendulum is fixed, while the other end can swing freely. $\theta$ denotes the angle of the pendulum from vertical position. The state contains the angle and its time derivative, i.e., $\ubm = (\theta, \dot{\theta})$. In each episode, the pendulum starts from a random state $\ubm_0 \in (-1,1)\times(-1, 1)$. Besides, the angular velocity is constrained as $\dot{\theta} \in [-8, 8]$, and any $\dot{\theta}$ out of this range will be capped by the boundary value ($-8$ or $8$). The dynamics of the pendulum system is governed by,
\begin{equation}
    \ddot{\theta} = -\frac{3 g}{2 l} \sin{(\theta + \pi)} + \frac{3}{m l^2}T, 
    \label{eq:pendulum}
\end{equation}
where $g = 10$ is the acceleration of gravity, $T \in [-2, 2]$ denotes the torque that the agent applies to the pendulum,  $l = 1$ and $m = 1$ are the length and mass of the pendulum, respectively. The control goal here is to swing the pendulum up and make it stays upright, meanwhile consuming as less energy as possible. To achieve this, the reward function is defined as,
\begin{equation}
    r = - \theta ^2 - 0.1 \dot{\theta}^2 - 0.001 T^2 
    \label{eq:PenRew}
\end{equation}

The performance curve is shown in Fig.\ref{fig:pen} (b). After about 3000 time steps, both the PiMBRL and MBRL achieve averaged total return of -200 with reduced uncertainty. In contrast, the total return of the MFRL largely fluctuates and the average value remains less than -800. Although PiMBRL shows greater sample efficiency over the MFRL baseline, it does not show a notable advantage over the MBRL counterpart in the Cart-Pole case.
This might be due to the following two factors. First, compared to the Cart-Pole problem, the dynamics of the pendulum environment is easier for the model to learn, since the observation space is much smaller than that of the Cart-Pole environment (2-dimension vs. 4-dimension). Second, the model network is benefited less from the physical loss, since when the state $\dot{\theta}$ exceeds the limits $[-8,8]$ and capped by the boundary value, the governing equation eq.(\ref{eq:pendulum}) is no longer satisfied. 

\begin{figure}[htp!]
    \centering
    \subfloat[Model trained with the physics-informed loss in PiMBRL]{\includegraphics[width=0.9\textwidth]{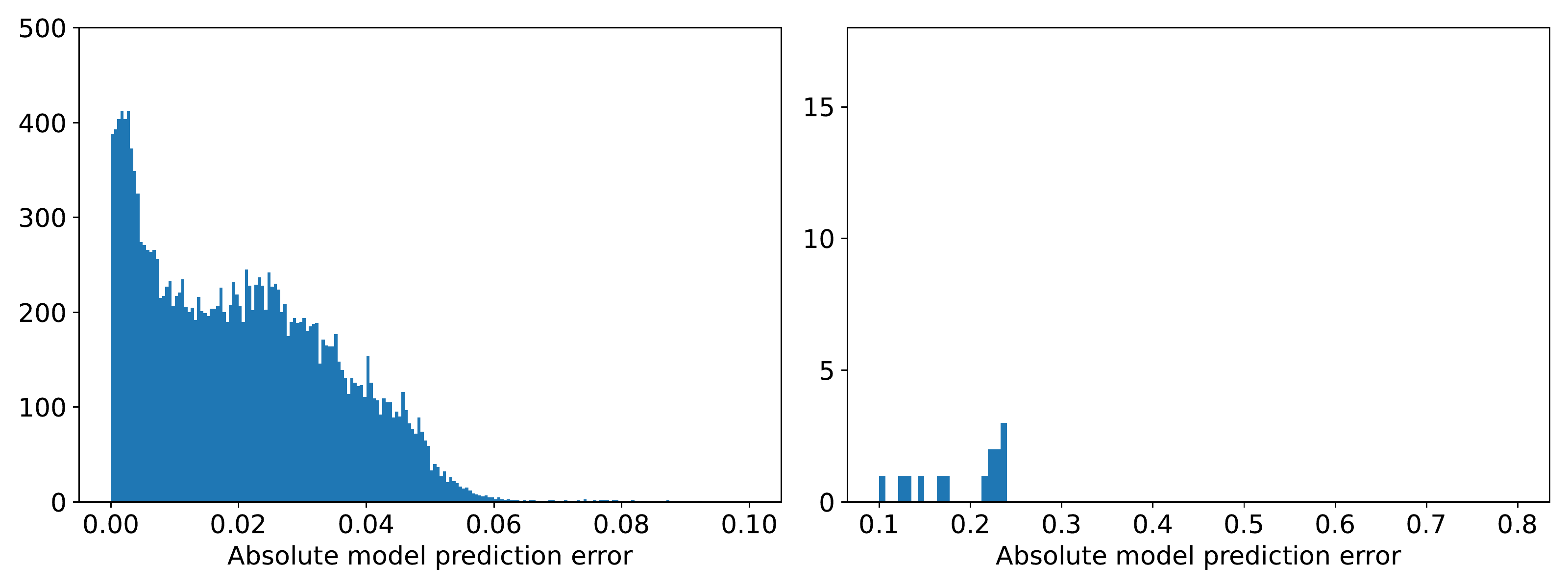}}\\
    \subfloat[Model trained with the data loss only in MBRL]{\includegraphics[width=0.9\textwidth]{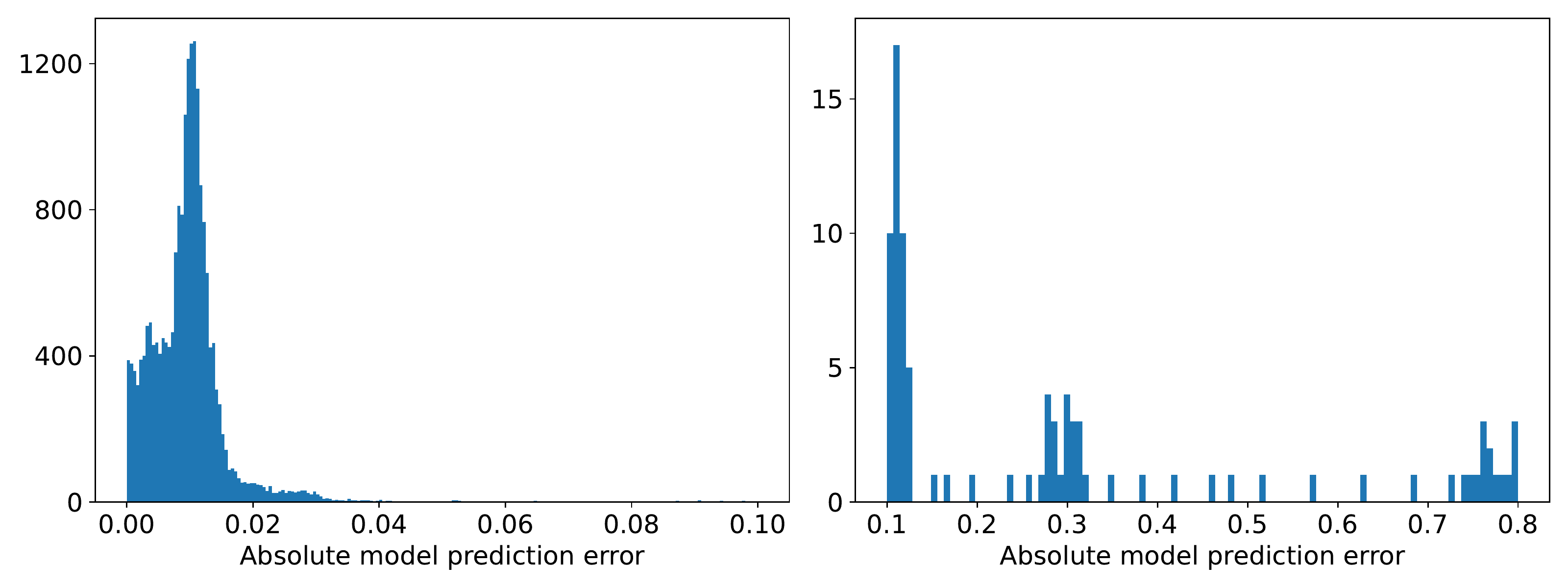}}
    \caption{Histograms of the prediction errors of the transition models for the Pendulum environment, (a) trained with physics-informed loss in PiMBRL, and (b) trained with the data loss only in MBRL.}
    \label{fig:pen_model}
\end{figure}
Figure~\ref{fig:pen_model} compares the histograms of the prediction errors of the models trained by PiMBRL and MBRL, respectively. Fig.~\ref{fig:pen_model} (a) shows the prediction error of the model trained with the physics-informed loss (equation loss + data loss), while Fig.~\ref{fig:pen_model} (b) shows the model prediction error in MBRL where only the labeled data are used for training. Although the model does achieve higher accuracy on average without using physics constraints (see two sub-figures in the left column), in the out-of-sample regime (away from the training set), the physics-informed model shows better performance and robustness (see two sub-figures in the right column). Overall, the models in both PiMBRL and MBRL are learned sufficiently well to achieve a roughly similar RL performance.   


\subsection{PDE governed environments}\label{sec:PDEs}
Unlike the environments that can be described by ODEs, the systems governed by PDEs are much more complicated in terms of the dimension of spatiotemporal solution space, dramatically increasing the level of difficulty in learning the dynamic model as well as the policy and value functions. In this section, we evaluate the proposed PiMBRL structure on two continuous control problems in the environments governed by Burgers' equation and Kuromoto-Sivashinsky (KS) equation, respectively. The learned transition model is only used to predict a limited rollout length ($l_M$) in one trajectory to control the model error accumulation, making it remain at a relatively low level. 

\subsubsection{Burgers' equation}
For the first PDE-based control problem, we consider a 1-D Burgers' equation with periodic boundary condition,  
\begin{equation}
    \frac{\partial u}{\partial t} + \frac{1}{2} u \frac{\partial u}{\partial x} = \nu \frac{\partial^2 u}{\partial x^2} + f(x,t), \quad x\in [0,l],\  t \in [0,2\pi],
    \label{eq:burgers}
\end{equation}
where $x$ is the spatial coordinate, $\nu = 0.01$ is the kinematic viscosity, and $f(x,t)$ denotes the source term, defined as,
\begin{equation}
    f(x,t) = a_1(t) \exp\left[{\left(-15(\frac{x}{l} - 0.25)\right)}^2\right] + a_2(t) \exp\left[{\left(-15(\frac{x}{l} - 0.75)\right)}^2\right]
    \label{eq:burgersF}
\end{equation}
with the control parameters $\abm=(a_1,a_2) \in {[-0.025,0.075]}^2$. 

The control problem is defined as training the RL agent to match a reference trajectory. Namely, the RL agent is trained to find the optimal strategy of controlling the source term with two control parameters $a_1,a_2$, in order to match a predefined reference trajectory profile $\ubm_{re}$, 
\begin{equation}
\label{eq:burgersRef}
    u_{re}(x,t) = 0.05 \sin{t} +0.5, \quad t \in [0,2\pi].
\end{equation}
Each episode starts from a randomly generated initial condition,
\begin{equation}
\label{eq:burgersIC}
    u(x,0) = 0.2\, c \exp \left[{(-5(\frac{x}{l}-0.5))}^2\right] + 0.2\, (1-c) \left(0.5 \sin{4\pi \frac{x}{l}}+0.5\right),
\end{equation}
where $c$ is randomly sampled from a uniform distribution on $[\,0,1)$. That is, the trained RL is expected to finally match the reference trajectory, starting from any randomly generated initial state by eq.(\ref{eq:burgersRef}). The observation is set as the discrepancy between the PDE state and reference state at the same control step $\ubm^o = \ubm - \ubm_{re}$.  The environment is simulated numerically based on the finite difference methods, where the convection term and diffusion term are discretized by the second-order upwind scheme and fourth-order central difference scheme, respectively. Euler method is used for the time integration. The simulated environment is defined on a spatial mesh of 150 grid points and the numerical time step is set as 0.01. The control signal is applied every 500 numerical steps and one episode contains 60 control steps. The reward function is defined as $-10\left\|\ubm^o\right\|_{L_2}$. 
\begin{figure}[htp!]
    \centering
    \subfloat[Uncontrolled]{\includegraphics[width=0.46\textwidth]{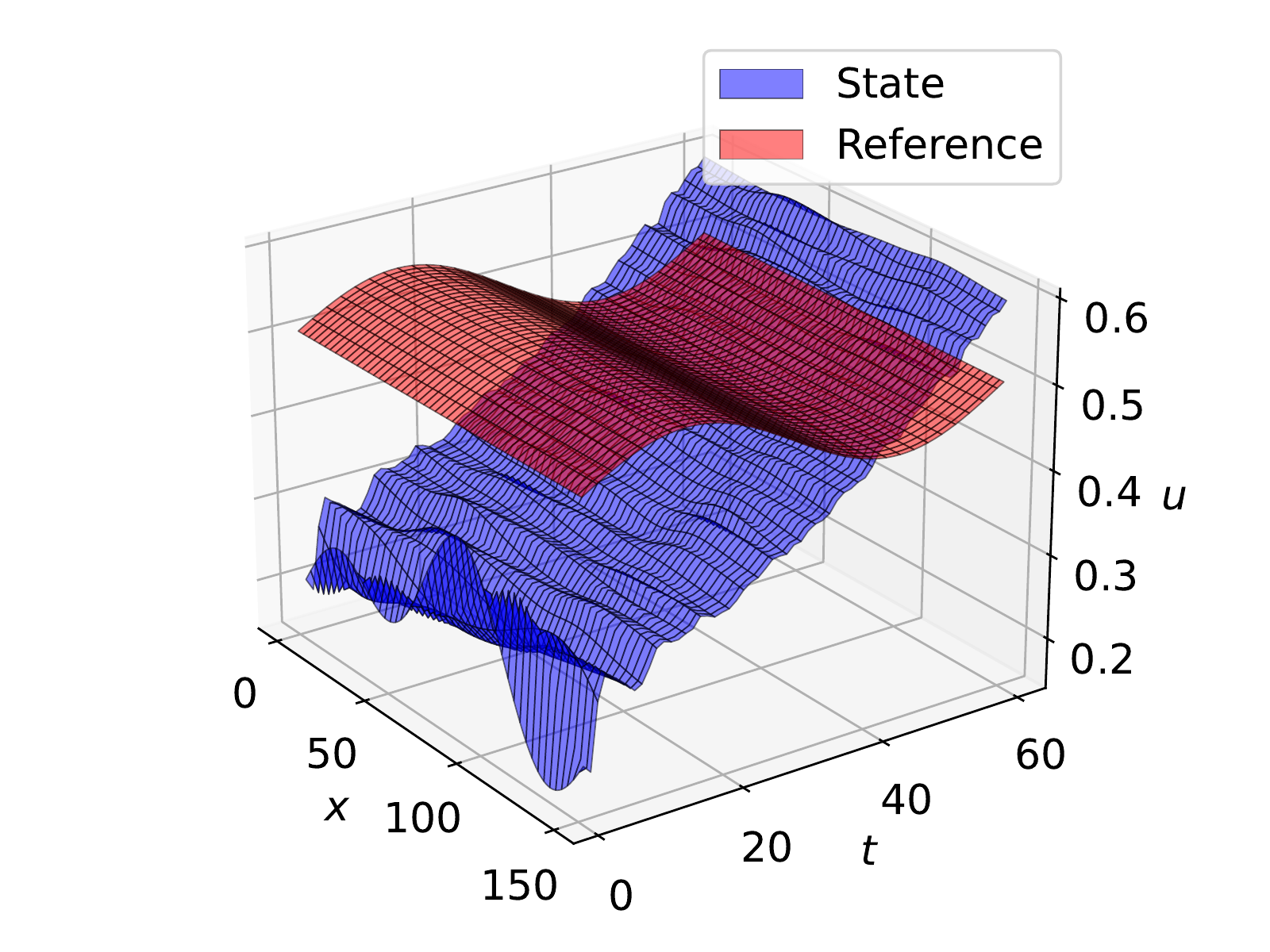}}
    \subfloat[Uncontrolled actions \& rewards]{\includegraphics[width=0.46\textwidth]{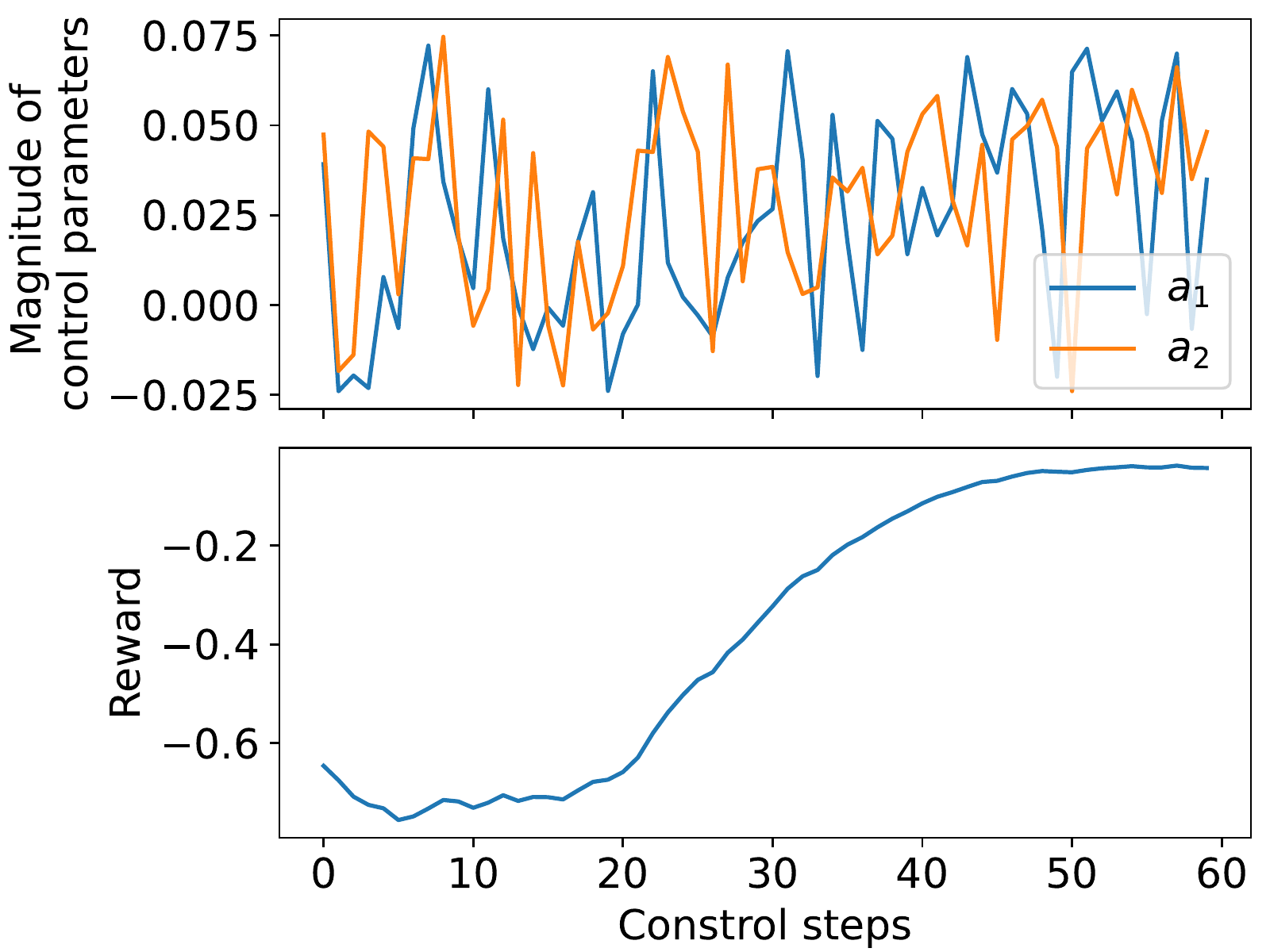}}\\
    \subfloat[Controlled]{\includegraphics[width=0.46\textwidth]{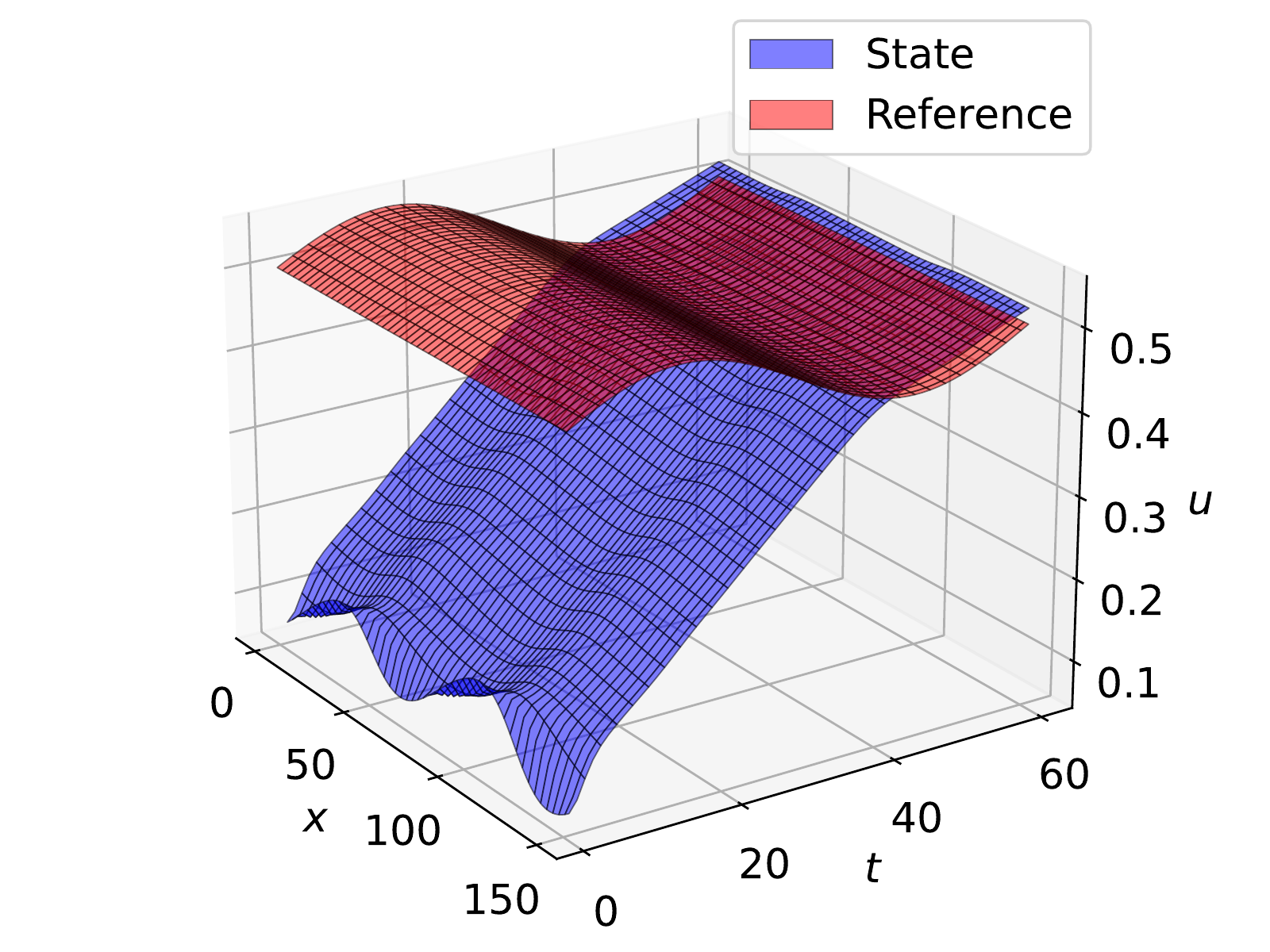}}
    \subfloat[Controlled actions \& rewards]{\includegraphics[width=0.46\textwidth]{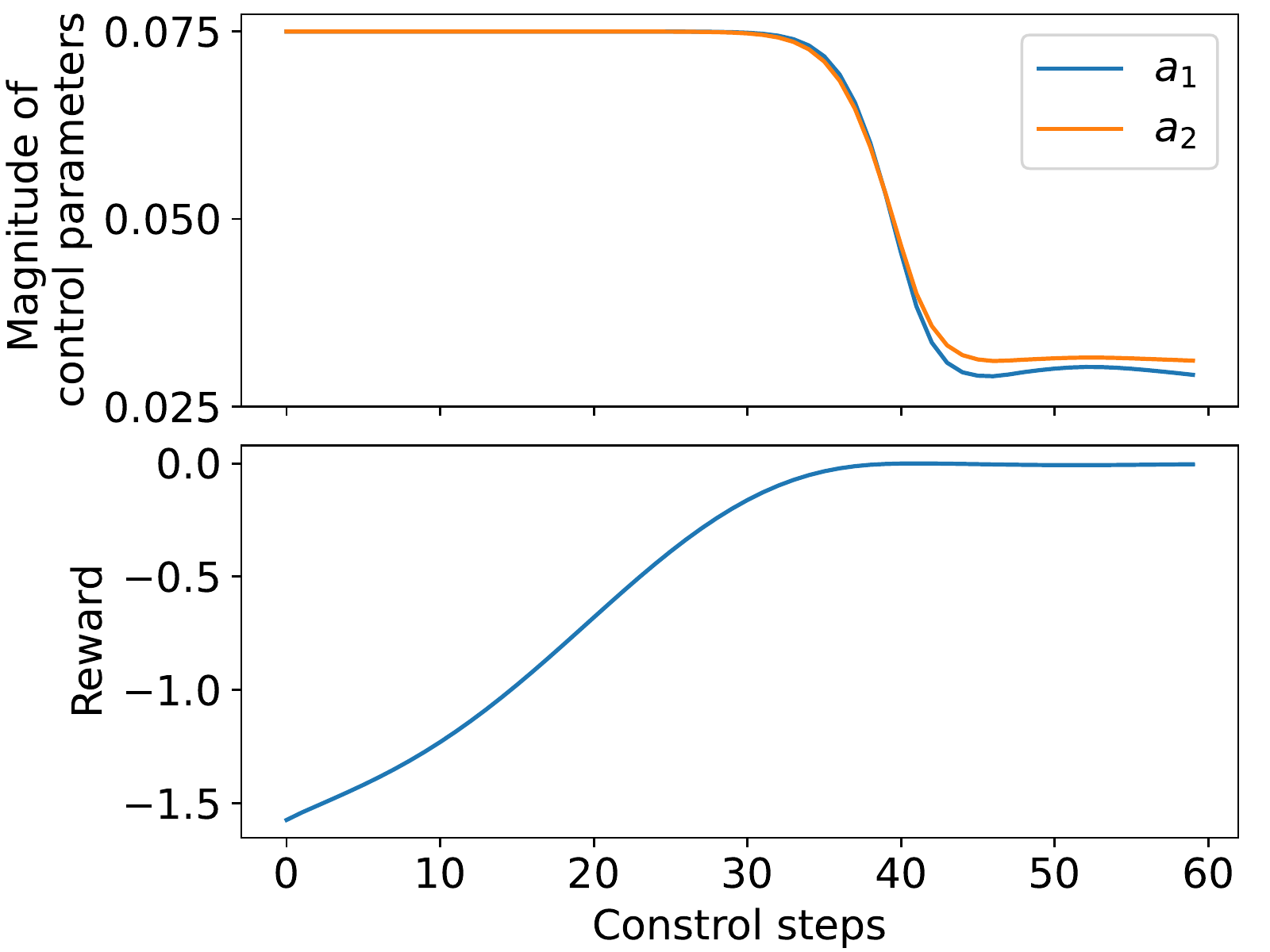}}
    \caption{Results of one test episode (a) with random control signals and (c) with trained RL controller. The corresponding actions and reward curves of the (b) uncontrolled episode and (d) RL controlled episode.}
    \label{fig:burgers_demo}
\end{figure}
Without the RL training, random control signals are applied to the system, and the corresponding spatiotemporal state surface $\ubm$ is shown in Fig.~\ref{fig:burgers_demo}a. We can see that the state surface is unsmooth, and a large discrepancy remains between the uncontrolled state with the reference state. Figure~\ref{fig:burgers_demo}c shows one of the test episodes controlled by the trained PiMBRL agent. The corresponding actions \& rewards are given in Fig.~\ref{fig:burgers_demo}d. Although the initial state is far from the reference, the controlled surface gradually approaches and finally matches the reference state after $t = 40$. The corresponding action curves are smooth, suggesting that the RL agent successfully learns an effective control strategy.



\begin{figure}[htp!]
    \centering
    \includegraphics[width=0.5\textwidth]{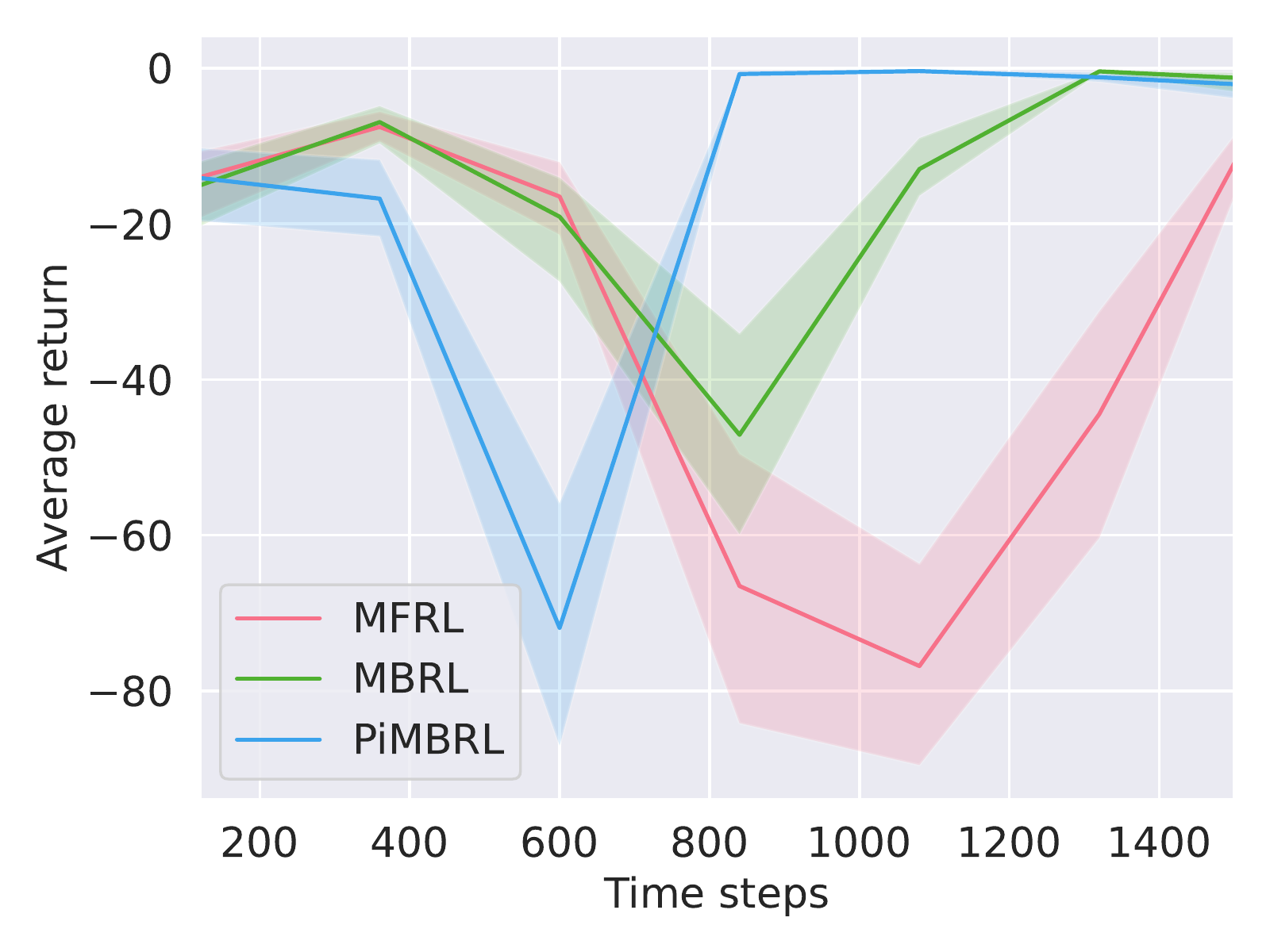}
    \caption{Performance curve of PiMBRL versus standard MFRL and MBRL in the Burgers' equation environment. The solid lines indicate averaged returns of 100 test episodes, while the shaded area represents the return distribution of all test samples.}
    \label{fig:burgers_perf}
\end{figure}
Figure~\ref{fig:burgers_perf} shows the performance curves of the PiMBRL, MBRL, MFRL tested on 100 randomly selected initial conditions. The PiMBRL reaches the total return of 0.1 only after 800 time steps, while it takes the MBRL about 1300 time steps to achieve a similar level of performance. The MFRL counterpart can not reach the same level of performance within 1400 time steps. Again, our PiMBRL shows significant advantages in terms of sample efficiency, since it only uses about 65\% of time steps required by MBRL and 46.7\% time steps required by MFRL to achieve the control goal.

\subsubsection{Kuramoto-Sivashinsky (KS) Equation}\label{sec:ks}
In the last case, we evaluate the proposed PiMBRL on the control of a nonlinear, chaotic dynamic system governed by the 1-D Kuramoto-Sivashinsky (KS) equation, which is more challenging. The system governed by the KS equation often exhibits spatiotemprally chaotic or weakly turbulent behavior, and thus the KS equation is widely used as a model system for turbulence study~\cite{cvitanovic2010state}. In this case, the KS environment is controlled by four actuators distributed equally in space to minimize the energy dissipation and total input power. The physics of this system is governed by the KS equation,
\begin{equation}
    \frac{\partial u}{\partial t} + \frac{\partial^2 u}{\partial x^2} + \frac{\partial^4 u}{\partial x^4} + \frac{1}{2} u\frac{\partial u}{\partial x} = f(x,t), \ \    x \in [0,l], \ t\in[0,\infty),
\end{equation}
where $u$ is the state variable, and $f$ represents the source term (i.e., actuator) defined by. 
\begin{equation}
    f(x,t) = \sum_{i=1}^{4}\frac{a_i(t) e^{-(x-x_i)^2/2}}{\sqrt{2\pi}},
\end{equation}
where $x_i \in \{0,\,l/4,\,l/2,\,3l/4\}$ is the spatial locations of the actuator, and $\abm = \{a_i(t)\}_{i=1,2,3,4} \in [-0.5,0.5]^4$ defines the control parameters. 
To achieve the control goal, the reward function is defined as follows,
\begin{equation}
    r = -\frac{1}{T l}\int_{t_0}^{t_0+T}\int_0^l \left((\frac{\partial^2 u}{\partial x^2})^2 +(\frac{\partial u}{\partial x})^2 + u f\right)\,dx\, dt 
\end{equation}
where $T$ is the time length of one control step. The environment is simulated numerically based on the finite difference method, where the convection term is discretized by the second-order upwind scheme, the second and fourth derivatives are discretized by the 6th-order central difference scheme, and the 4th-order Runge-Kutta scheme is used for time integration with time stepping size of 0.001 on the 1D domain $l=8\pi$ discretized by a mesh of 64 grid points.
\begin{figure}[htp!]
    \centering
    \includegraphics[width=0.7\textwidth,valign=b]{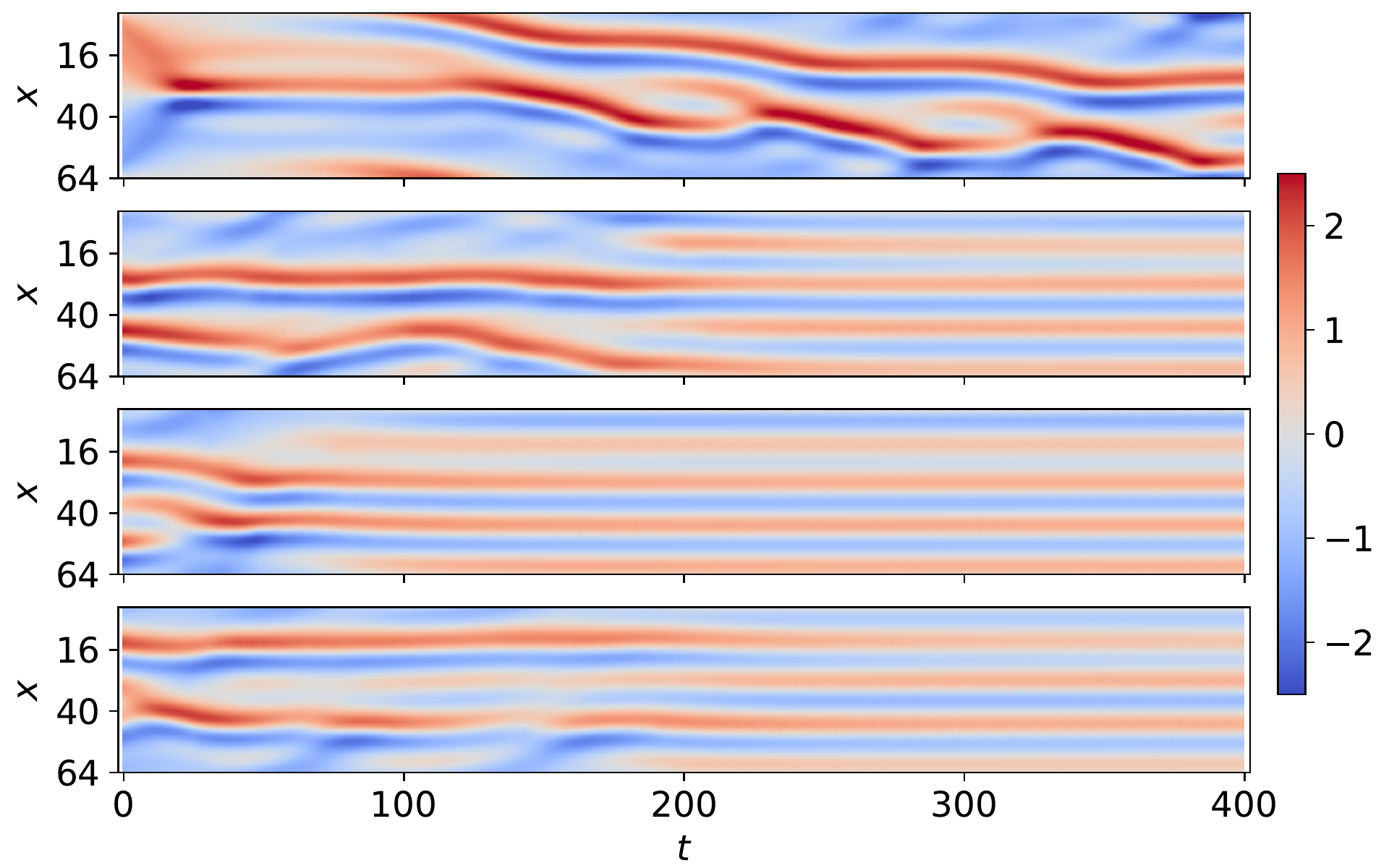}
    \caption{Contours of spatiotemporal state $u$ of four randomly-selected episodes. The top one is an uncontrolled episode, while the other three are controlled by the trained RL agent.}
    \label{fig:ks_state}
\end{figure}
Each control step contains 250 numerical steps,  and one episode consists of 400 control steps. Each episode starts with a random initial condition sampled from the attractor of the unforced KS equation. Figure~\ref{fig:ks_state} shows the spatiotemporal states of four test episodes with randomly sampled initial states. The top one is an uncontrolled episode, where nonlinear chaotic behavior is developed along the time axis. In contrast, the ``turbulence'' in the other three episodes controlled by the agent can be quickly stabilized after 200 time steps, showing the effectiveness of the PiMBRL controller.

\begin{figure}[htp!]
    \centering
    \includegraphics[width=0.55\textwidth,valign=c]{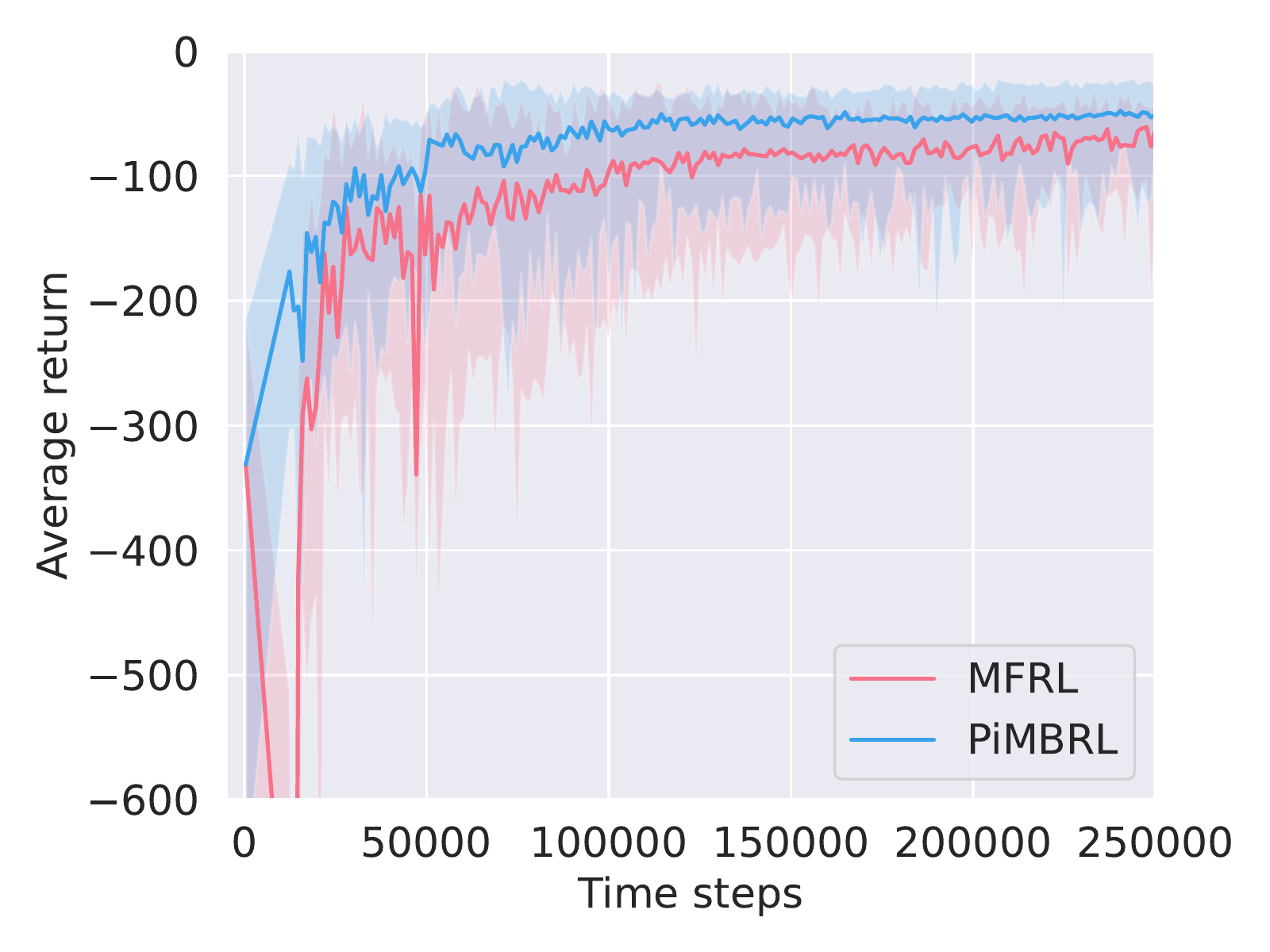}
    \caption{Performance curve of PiMBRL versus MFRL and MBRL in KS equation environment. Solid lines shows the average return of all the test episodes while the shaded area represents the distribution range of the 200 test episodes}
    \label{fig:ks_perf}
\end{figure}

Figure~\ref{fig:ks_perf} shows the performance curve of PiMBRL versus that of the MFRL. It is clear that the PiMBRL agent reaches higher averaged total returns with fewer time steps than the MFRL counterpart does. The PiMBRL performance curve is consistently above that of the MFRL approach, and meanwhile, less uncertainty is observed. In PiMBRL, the agent only uses about $30.2\%$ of the time steps required by its MFRL counterpart to reach the averaged total return of $-55$. In this case, a model-free fine-tuning approach~\cite{nagabandi2018neural} is applied when the averaged return is above -55 to further improve the PiMBRL performance. This is because when the RL agent is trained to achieve a high accuracy level, the model-based exploration does not help too much while the model bias becomes the bottleneck. At this point, the RL agent can be fine-tuned by interacting with the real environment without using the model~\cite{nagabandi2018neural}. As such, the PiMBRL and MFRL performance curves will eventually converge to each other with a large number of training time steps.

\begin{figure}[H]
    \centering
    \includegraphics[width=0.9\textwidth]{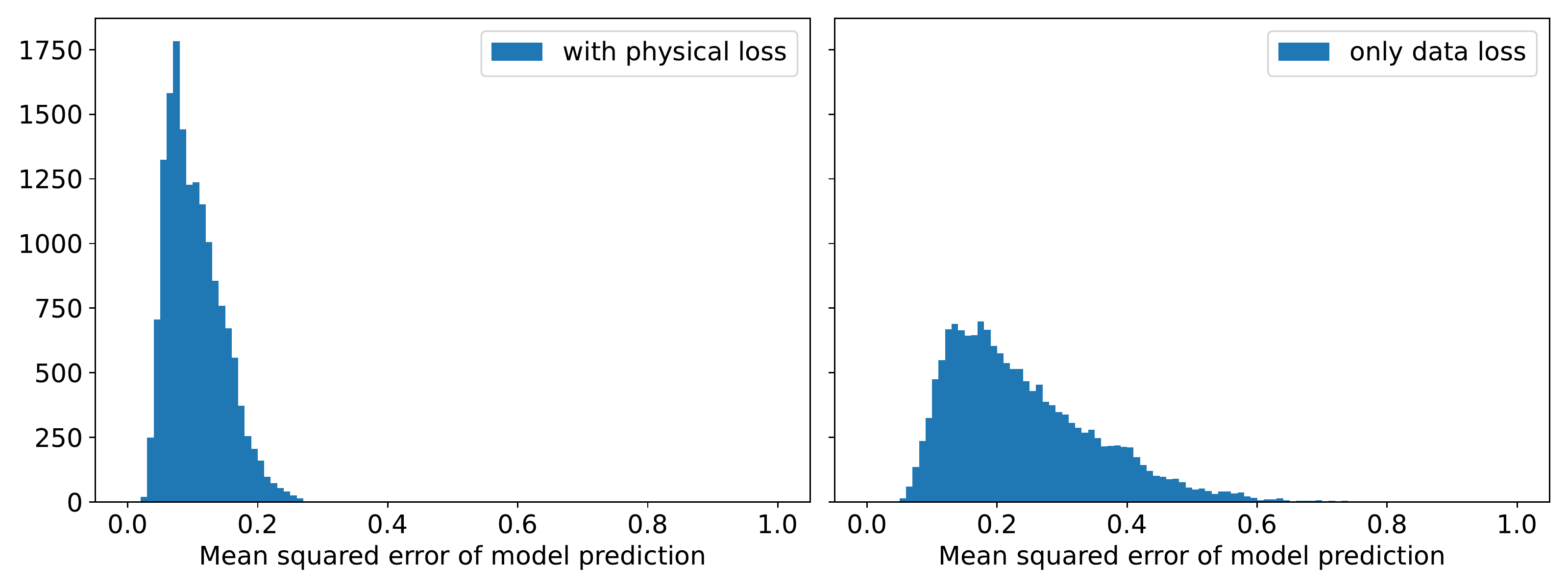}
    \caption{Histograms of the prediction errors (MSE) of the transition models for the KS system, (left) trained with physics-informed loss in PiMBRL, and (right) trained with the data loss only in MBRL. Both are trained on the same real buffer with $10^5$ time steps stored.}
    \label{fig:ks_model_error}
\end{figure}

The MBRL result is not plotted here because the accuracy ($L_D$) of the learned model purely based on labeled data does not meet the accuracy threshold $\lambda = 0.01$, and thus the MBRL performance curve is identical to that of the MFRL counterpart. In contrast to the ODE-based pendulum system shown above, the spatiotemporal dynamics of the chaotic system is much more challenging for the model to learn purely based on the limited amount of labeled data. Incorporating physics constraints can significantly help this scenario and improve the model learning performance. This can be clearly seen in Fig.~\ref{fig:ks_model_error}, where the comparison of model prediction errors of PiMBRL (left) and MBRL (right) are shown. The mean squared error distribution of the model trained based on the physics-informed loss is much less than that of the model trained purely based on the labeled data.


\section{Discussion}
\label{sec:discuss}

\subsection{Influence of model rollout length}\label{sec:modellength}
The rollout length of the model is important to the RL performance. On the one hand, a long-term model prediction could help the agent see ``deeper and further'', improving the exploration rate and increasing the sample efficiency in the Dyna-like MBRL framework. On the other hand, accurately predicting a long trajectory is always challenging due to the error accumulation effect, and inaccurate prediction data could be harmful to the RL training, which is a trade-off.  
\begin{figure}[htp!]
    \centering
    \includegraphics[width=0.8\textwidth]{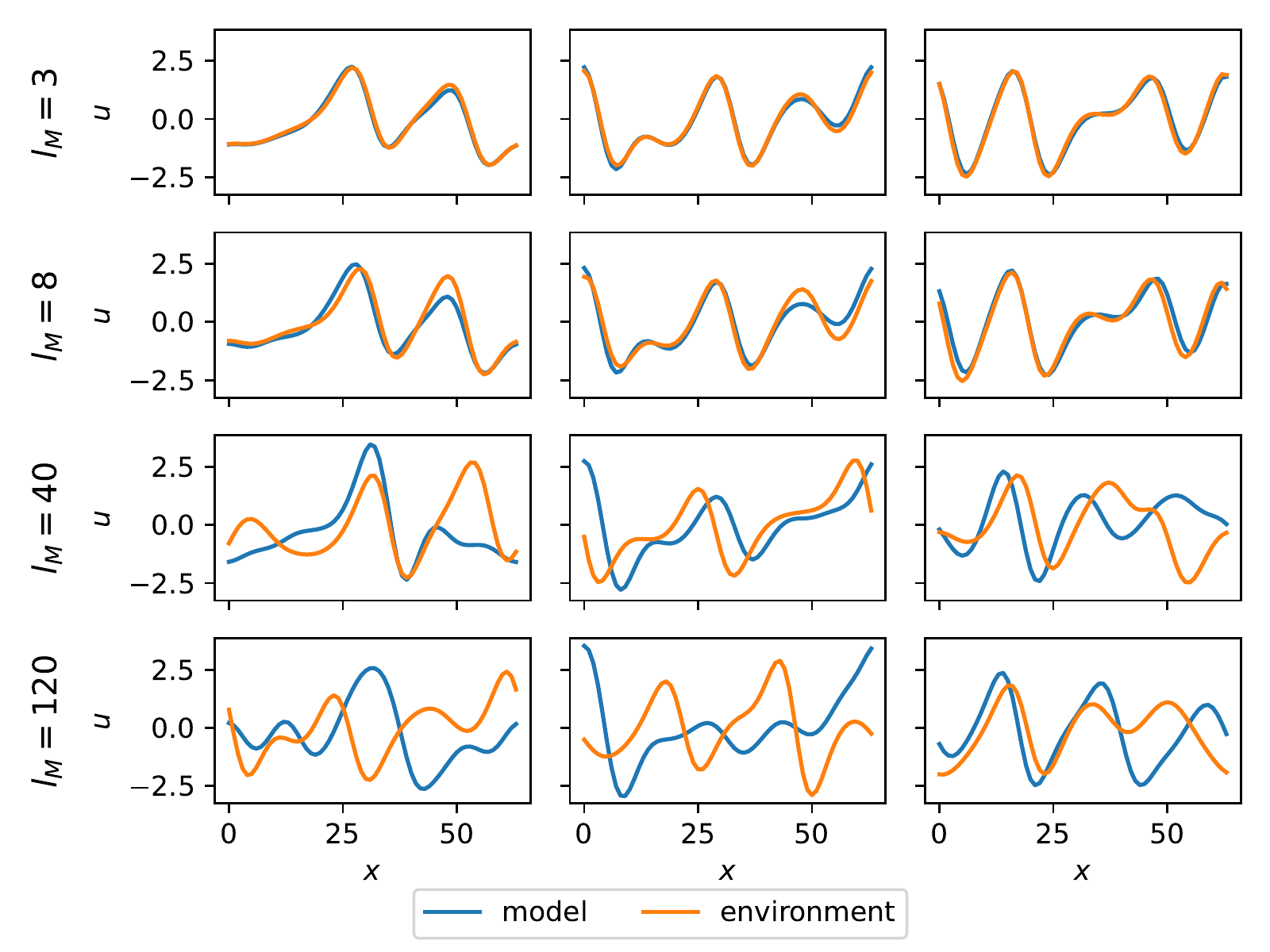}
    \caption{Model predicted snapshots of the KS environment with different rollout lengths compared with the true environment at three randomly selected initials.}
    \label{fig:dis_ks_model}
\end{figure}
Figure~\ref{fig:dis_ks_model} shows the model prediction performance for the KS environment with four different rollout lengths ($l_M = 3, 8, 40, 120$). For all three randomly selected initials, we can see that the states predicted by the model agree well with the ground truth when the rollout length is small ($l_M <= 8$). However, large discrepancies can be observed when $l_M >= 40$ and the model predictions significantly deviate from the ground truth when the rollout length is 120. 


We are more interested in how the model rollout length affects the PiMBRL performance, so we investigated the influence of four different rollout lengths on the reward curves. The case setting is the same as that used in section \ref{sec:result}\ref{sec:PDEs}\ref{sec:ks} except for the rollout length, and the RL performance is evaluated on 200 randomly generated episodes. 
\begin{figure}[htp!]
    \centering
    \includegraphics[width=0.52\textwidth]{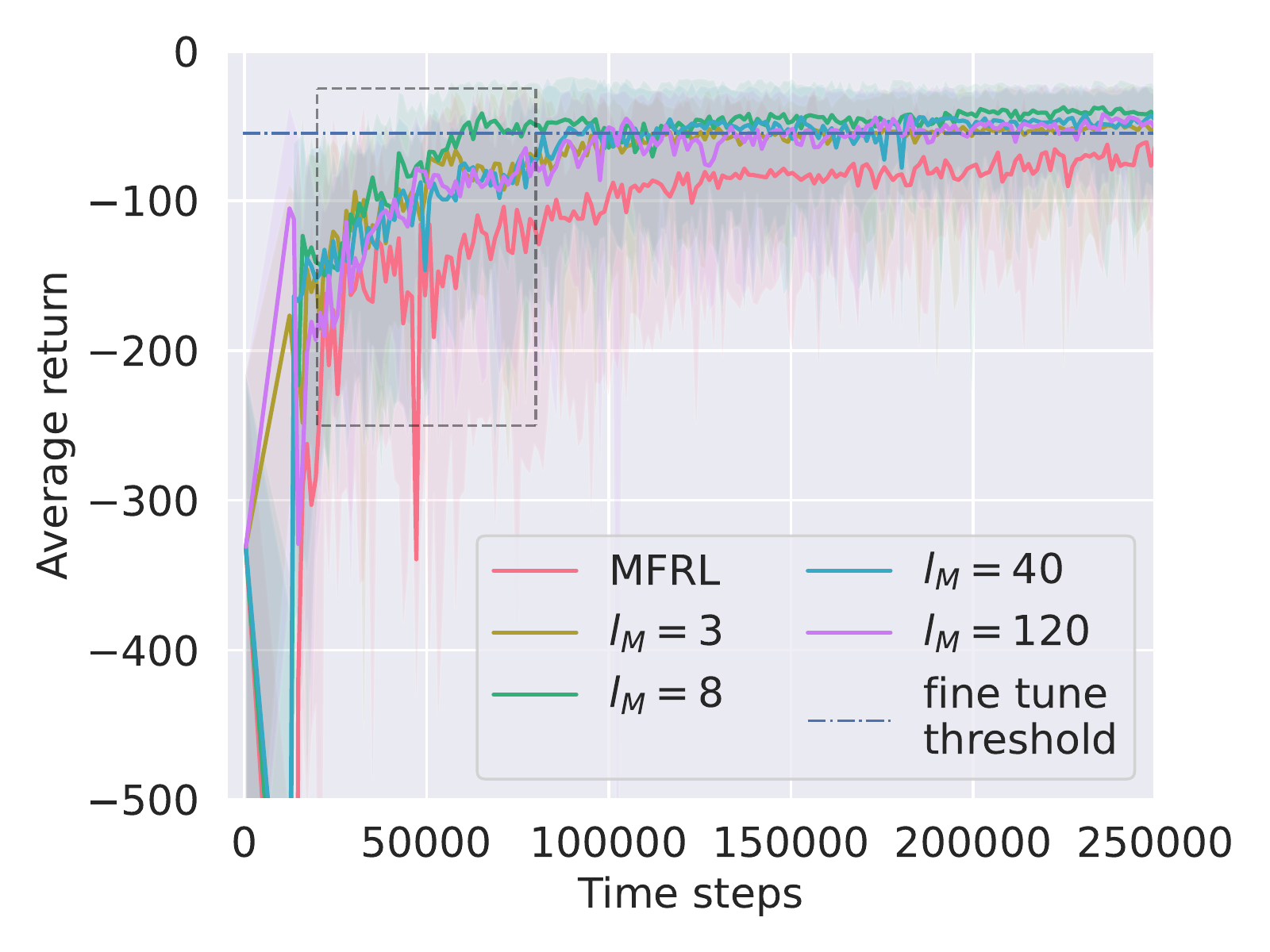}\\
    \includegraphics[width=0.48\textwidth]{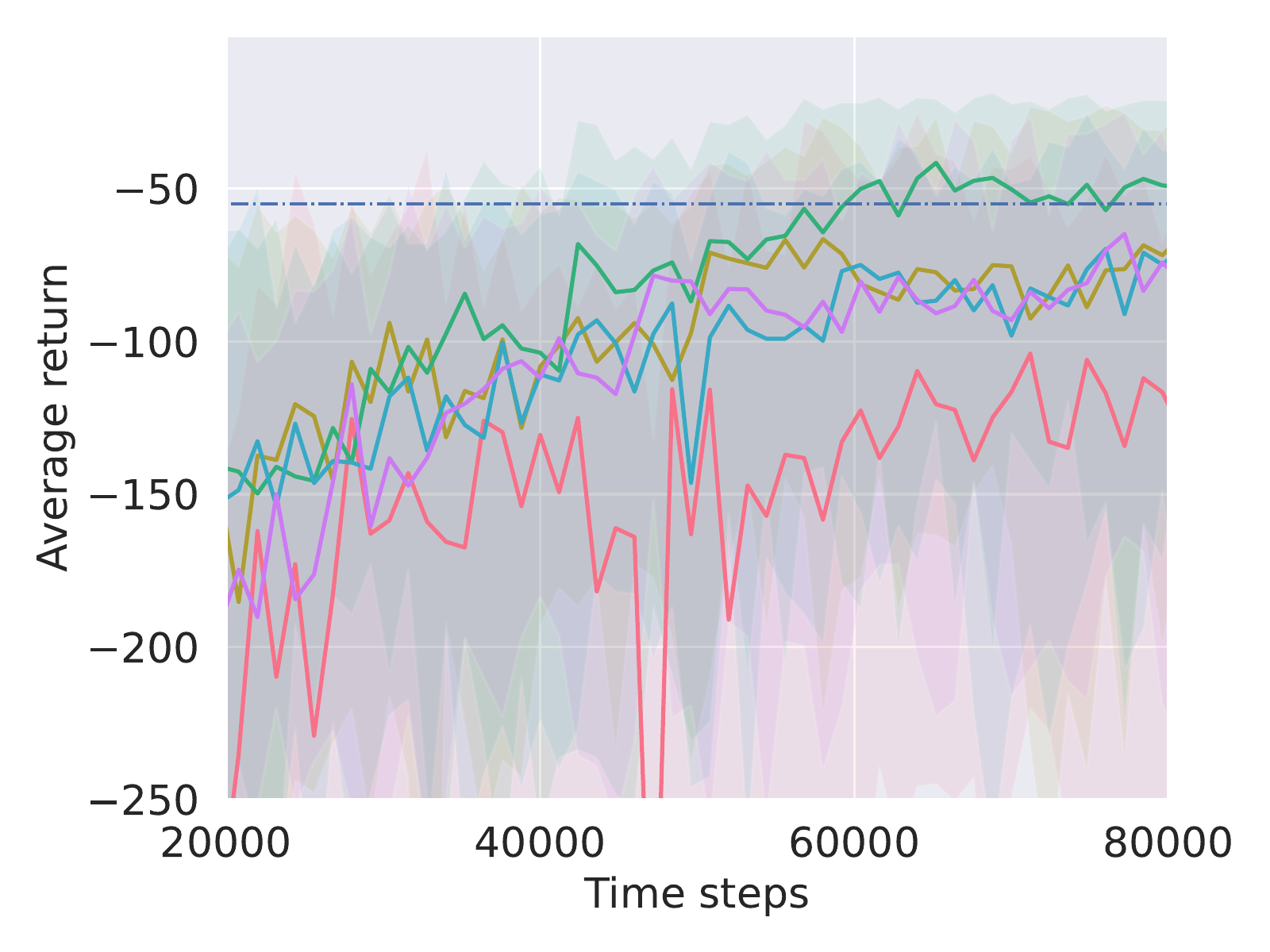}
    \includegraphics[width=0.48\textwidth]{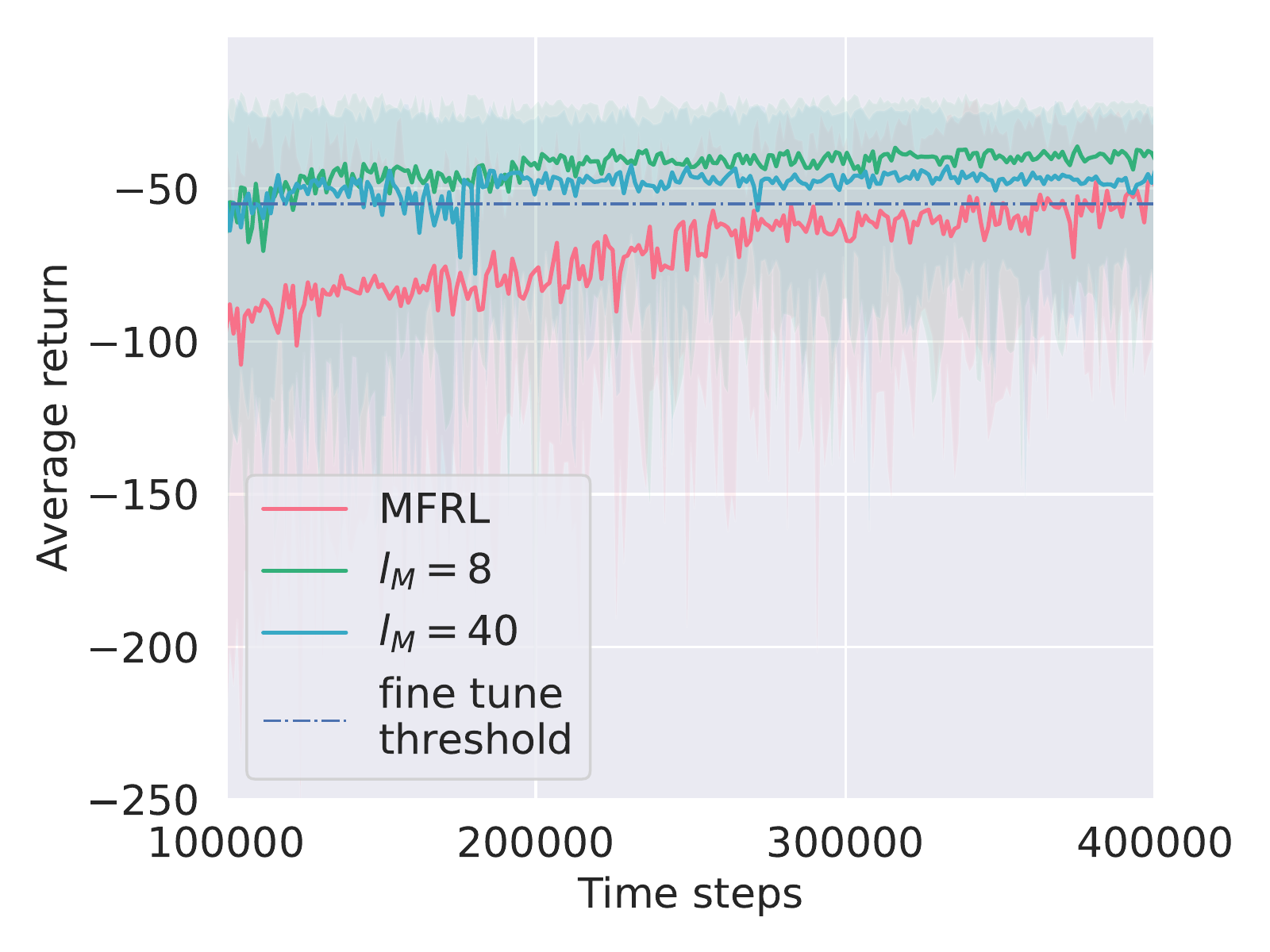}
    \caption{Performance curves of PiMBRL with four different model rollout lengths compared with MFRL base line. (Left bottom) Zoom-in view of the dashed box. (right bottom) Zoom-in view of the range between $1\times10^5$ to $4\times10^5$ time steps.}
    \label{fig:dis_ks_pl}
\end{figure}
Figure~\ref{fig:dis_ks_pl} shows the performance curves of PiMBRL with different model prediction lengths. As expected, the RL performance slightly deteriorates if the rollout length of the model is either too short or too long. For the four rollout lengths ($l_M = 3, 8, 40, 120$), the RL agent achieves the best performance with $l_M = 8$ at almost any stage of the entire training process. When $l_M <= 8$ or $l_M >=8$, the RL convergence speed is relatively slow before entering the fine-tuning stage due to the trade-off between exploration and model bias. However, even after the model-free fine tuning, the RL agent with a longer rollout length ($l_M >= 40$) still suffers from the low-quality model prediction data and is very difficult to be further improved (see the comparison of the green and blue curves in Fig.~\ref{fig:dis_ks_pl}).


\subsection{Influence of model accuracy threshold.}
In model-based RL, the transition model is trained along with the RL agent (i.e., value and policy networks) from scratch, and the model-generated data usually can be effectively used once the model is trained to reach a certain level of accuracy. As mentioned above, we use the model accuracy threshold parameter ($\lambda$) to determine when the model-predicted data should be utilized for the RL training. Here, we would like to study how this parameter of $\lambda$ affects the PiMBRL performance.    

\begin{figure}[htp!]
    \centering
    \subfloat[Full view]{\includegraphics[width=0.5\textwidth]{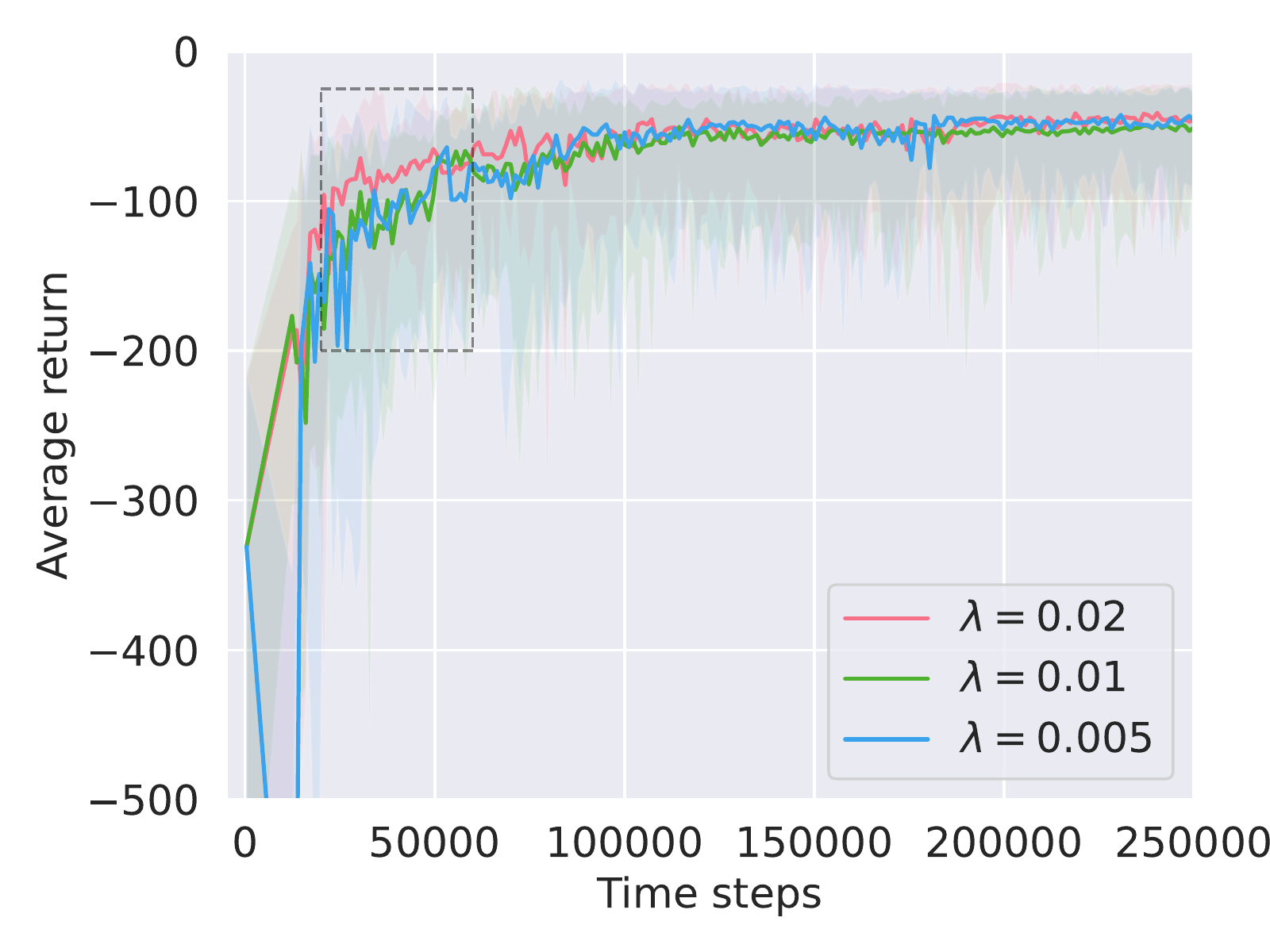}}
    \subfloat[Zoom-in view]{\includegraphics[width=0.5\textwidth]{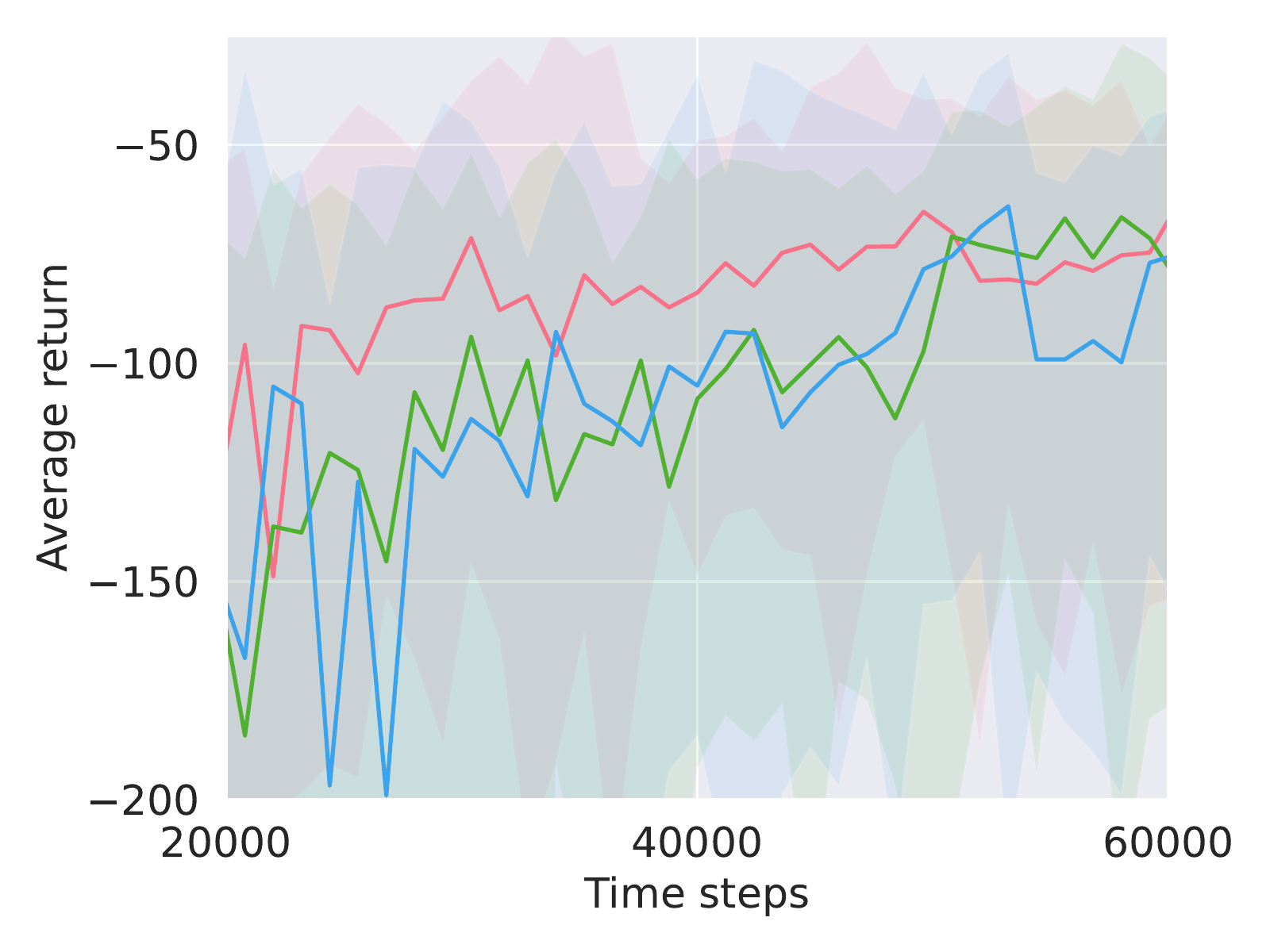}}
    \caption{Performance curves of PiMBRL with three different model accuracy threshold values $\lambda = 0.02, 0.01, 0.005$, where (a) is the full view of the entire training history and (b) shows the zoom-in view of the dashed box area in (a).}
    \label{fig:dis_ks_threshold}
\end{figure}

Using the KS environment as an example, Fig.~\ref{fig:dis_ks_threshold} shows the performance curves of PiMBRL with three different threshold values, i.e., $\lambda = 0.02, 0.01, 0.005$. Overall, the influence of $\lambda$ values on the final performance of the RL agent is negligible since all cases converge to the same level of total return. This is because the model is trained together with the RL agent and can be improved over the entire RL training process. Actually, with the same amount of training time steps in the real environment, the models can reach a similar accuracy level regardless of threshold values $\lambda$. However, the RL agent with a relatively large threshold value ($\lambda = 0.02$) performs slightly better and has a relatively faster convergence rate at the early stage of the training (time step 40000 to 60000). A higher threshold allows the agent to access more data generated by the model earlier, which leads to a better exploration rate and thus slightly higher sample efficiency (see Fig. \ref{fig:dis_ks_threshold}b).  

\section{Conclusion}
\label{sec:conclude}

In this work, we presented an innovative model-based reinforcement learning framework (PiMBRL) for dynamic control, which leverages the physical laws and constraints of the environment to alleviate the model-bias issue and significantly improve the sample efficiency. In particular, an autoencoding based recurrent network structure is devised to learn the spatiotemporal dynamics of the environments. The merit and effectiveness of the proposed PiMBRL framework have been demonstrated over a set of classic dynamic control problems, where the environments are governed by canonical ODEs or PDEs, including viscous Burgers' equations and KS equations with chaotic behaviors. Compared to the model-free or purely data-driven model-based reinforcement learning counterparts, our PiMBRL shows a significant improvement in model predictive accuracy and RL sample efficiency. Moreover, the effects of different hyper-parameters used in PiMBRL (e.g., model rollout length $l_M$ and model accuracy threshold $\lambda$) are studied, and how these parameters affect the RL performance is discussed.


\enlargethispage{20pt}


\dataccess{The data that support the findings of this study will be openly available in GitHub at https://github.com/Jianxun-Wang/PIMBRL upon publication.}

\aucontribute{X.-Y. Liu: participated in the design of the study, implemented the entire framework, carried out all numerical experiments, and drafted the manuscript. J.-X. Wang: conceived of the study, drafted and revised the manuscript, and supervised the project.}

\competing{We declare we have no competing interests.}

\funding{This work is funded by the National Science Foundation under award numbers CMMI-1934300 and OAC-2047127 and startup funds from the College of Engineering at University of Notre Dame.}



\section*{Appendix}\label{sec:appendix}
\subsection{Twin-delayed deep deterministic policy gradient (TD3) }

\begin{algorithm}[htp!]
\caption{Model-free Twin-delayed deep deterministic policy gradient (TD3)}
\begin{algorithmic}[1]
\State Initialize policy (actor) network $\pi(\ubm;\boldsymbol{\theta}_{\pi})$, value (critic) networks $q_1(\ubm, \abm;\boldsymbol{\theta}_{q_1})$, $q_2(\ubm, \abm;\boldsymbol{\theta}_{q_2})$, empty the replay buffer $\mathscr{D}^r$ and reset the environment $\mathcal{F}$.
\State Make a copy of policy and value networks as target networks $\pi_{targ}\gets\pi$, $q_{targ, 1}\gets q_1$, $q_{targ, 2}\gets q_2$
\For{time steps $t=1, N$}
    \State Execute action $\abm_i = \pi(\ubm;\boldsymbol{\theta}_{\pi})$ in the environment $\mathcal{F}$;
    \State Save new data pair $(\ubm^o_i,\abm_i,\ubm^o_{i+1},r_{i},d_{i})$ to buffer $\mathscr{D}^r$;
    \If {episode ends}
        Reset the environment $\mathcal{F}$;
    \EndIf
    \If{$t \mod \texttt{update\_every} ==0$}
        \If{enough state-action pairs stored in $\{\mathscr{D}^r,\mathscr{D}^f\}$}
            \For{$k = 1, I_{RL}$}
                \State Sample $J$ state-action pairs $\{(\ubm^o_j,\abm_j,\ubm^o_{j+1}, r_j, d_j)\}$ from buffer $\mathscr{D}^r$
                \State Compute target 
                \begin{equation*}
                    Q_j = r_j + \gamma(1-d_j)\min_{i=1,2}\left\{q_{targ, i} (\ubm^o_{j+1}, \pi_{targ}(\ubm^o_{j+1}))\right\}
                \end{equation*}
                \State Update value networks via gradient descent,
                \begin{equation*}
                    \nabla_{\boldsymbol{\theta}_{q_i}}\frac{1}{J} \sum_{j=1}^J \left[q_i(\ubm^o_j, \abm_j) - Q_j\right]^2, \qquad i = 1, 2
                \end{equation*}
                \If{$k \mod 2 == 0$}
                    \State Update policy network via gradient descent:
                    \begin{equation*}
                        \nabla_{\boldsymbol{\theta}_{\pi}} \frac{1}{J} \sum_{j = 1}^J q_1 (\ubm^o_j, \pi(\ubm^o_j))
                    \end{equation*}
                    \State Update target networks:
                    \begin{equation*}
                    \begin{split}
                        \boldsymbol{\theta}_{q_{targ, i}} \gets \rho \boldsymbol{\theta}_{q_{targ, i}} + (1-\rho)\boldsymbol{\theta}_{q_i} ,\qquad i = 1, 2\\
                        \boldsymbol{\theta}_{\pi_{targ}} \gets \rho\boldsymbol{\theta}{\pi_{targ}} + (1-\rho)\boldsymbol{\theta}{\pi}\qquad\qquad\qquad
                    \end{split}
                    \end{equation*}
                \EndIf
            \EndFor
        \EndIf
    \EndIf
\EndFor

\end{algorithmic}\label{MF-TD3}
\end{algorithm}

TD3 hyper-parameters used in this work is shown in Table~\ref{tab:TD3-hp},
\begin{table}[H]
    \centering
    \begin{tabular}{|c|c|c|c|c|c|}
        \hline 
        Parameter & discount rate $\gamma$ & $P_D$ & $I_{RL}$ & \begin{tabular}{@{}c@{}} Policy / Value \\ network optimizer  \end{tabular} & \begin{tabular}{@{}c@{}} Policy / Value \\ network learning rate  \end{tabular}\\ 
        \hline
        Value & \begin{tabular}{@{}c@{}} 0.99 \\(0.977 in KS) \end{tabular} & 2 & 50 & Adam & 1e-3\\
        \hline
    \end{tabular}
    \caption{TD3 hyper-parameters used in this work}
    \label{tab:TD3-hp}
\end{table}

\begin{algorithm}[htp!]
\caption{TD3-based policy and value network update algorithm used in MBRL/PiMBRL}
\begin{algorithmic}[1]
\State With buffer $\mathscr{D}^r$, and $\mathscr{D}^f$, policy network $\pi(\ubm;\boldsymbol{\theta}_{\pi})$, value networks $q_1(\ubm, \abm;\boldsymbol{\theta}_{q_1})$, $q_2(\ubm, \abm;\boldsymbol{\theta}_{q_2})$ and their corresponding target networks
$\pi_{targ}, q_{targ, 1}, q_{targ, 2}$
\For{$k = 1, I_{RL}$}
            \State Sample $J$ state-action pairs $\{(\ubm^o_j,\abm_j,\ubm^o_{j+1}, r_j, d_j)\}$ from buffer $\{\mathscr{D}^r, \mathscr{D}^f\}$
            \State Compute target 
            \begin{equation*}
                Q_j = r_j + \gamma(1-d_j)\min_{i=1,2}\left\{q_{targ, i} (\ubm^o_{j+1}, \pi_{targ}(\ubm^o_{j+1}))\right\}
            \end{equation*}
            \State Update value networks by gradient descent:
            \begin{equation*}
                \nabla_{\boldsymbol{\theta}_{q_i}}\frac{1}{J} \sum_{j=1}^J \left[q_i(\ubm^o_j, \abm_j) - Q_j\right]^2, \qquad i = 1, 2
            \end{equation*}
            \If{$k$ mod 2 == 0}
            \State Update policy network by gradient descent:
            \begin{equation*}
                \nabla_{\boldsymbol{\theta}_{\pi}} \frac{1}{J} \sum_{j = 1}^J q_1 (\ubm^o_j, \pi(\ubm^o_j))
            \end{equation*}
            \State Update target networks:
            \begin{equation*}
            \begin{split}
                \boldsymbol{\theta}_{q_{targ, i}} \gets \rho \boldsymbol{\theta}_{q_{targ, i}} + (1-\rho)\boldsymbol{\theta}_{q_i} ,\qquad i = 1, 2\\
                \boldsymbol{\theta}_{\pi_{targ}} \gets \rho\boldsymbol{\theta}{\pi_{targ}} + (1-\rho)\boldsymbol{\theta}{\pi}\qquad\qquad\qquad
            \end{split}
            \end{equation*}
            \EndIf
        \EndFor
\end{algorithmic}\label{integ-TD3}
\end{algorithm}

Hyper-parameters of PiMBRL used in section \ref{sec:result} is summarized in Table~\ref{tab:MBRL_hp}.
\begin{table}[htp!]
    \centering
    \begin{tabular}{|c|c|c|c|c|c|}
         \hline
         Environment   & 
         \begin{tabular}{@{}c@{}} Model accuracy\\threshold $L_D$ \end{tabular}  & \begin{tabular}{@{}c@{}} Model rollout \\length $l_M$ \end{tabular}&\begin{tabular}{@{}c@{}} Maximum  \\episode length \end{tabular} & $n_{s_M}$ & $n_{s_R}$\\
         \hline
         Cart-Pole & $1$e$-4$ & $200$ (Full length)& $200$ & 800 & 1000\\
         \hline
         Pendulum & $1$e$-2$ & $200$ (Full length)& 200 & 6000 & 12000\\
         \hline
         Burgers' & $1$e$-2$ &  $1$ & 60 & 120 & 120\\
         \hline
         KS  & $1$e$-2$ & $3$ & 400 & 6000 & 12000\\
         \hline
    \end{tabular}
    \caption{Hyper-parameters of PiMBRL used in this work (section \ref{sec:result})}
    \label{tab:MBRL_hp}
\end{table}
\newpage
\clearpage

\bibliographystyle{unsrtnat}
\bibliography{./reference.bib}

\end{document}

%% file: preamble.macros.tex
\usepackage{bm} 
\usepackage{amsfonts} 

\newcommand\abm{{\ensuremath{\bm{a}}}}

\newcommand\ubm{{\ensuremath{\bm{u}}}}

\newcommand\xbm{{\ensuremath{\bm{x}}}}

\newcommand\mubold{{\ensuremath{\boldsymbol{\mu}}}}

